# D'RespNeT: A UAV Dataset and YOLOv8-DRN Model for Aerial Instance Segmentation of Building Access Points for Post-Earthquake Search-and-Rescue Missions

Aykut Sirma, Angelos Plastropoulos, Gilbert Tang, Argyrios Zolotas[a,b,*]

[a]Cranfield University, College Road, Milton Keynes, MK43 0AL, Bedfordshire, United Kingdom

[b]Centre for Autonomous and Cyberphysical Systems, Cranfield University, Milton Keynes, Bedfordshire, United Kingdom

**Abstract**

Recent advancements in computer vision and deep learning have significantly enhanced disaster-response capabilities, particularly in the rapid assessment of earthquake-affected urban environments. Timely identification of accessible entry points and structural obstacles is essential for effective search-and-rescue (SAR) operations, potentially saving lives in critical scenarios. To address this urgent need, this study introduces **D'RespNeT**, a novel high-resolution dataset specifically developed for aerial instance segmentation of post-earthquake structural environments. Unlike existing datasets, which rely heavily on satellite imagery or simplistic semantic labeling, **D'RespNeT** provides detailed polygon-level instance segmentation annotations derived from high-definition (1080p) aerial footage captured in recent disaster zones, including the 2023 Türkiye earthquake and other earthquake-impacted regions. The dataset comprises 28 operationally critical classes, including structurally compromised buildings, clearly identified access points (doors, windows, gaps), various debris levels, rescue personnel, vehicles, and civilian visibility. A distinctive innovation of **D'RespNeT** is its fine-grained annotation detail, enabling precise differentiation between accessible and obstructed areas, thereby substantially improving operational planning and response efficiency. Performance evaluations utilizing advanced YOLO-based instance segmentation models, specifically YOLOv8-seg, reveal significant improvements in real-time situational awareness and navigational decision-making. Compared to baseline approaches, our optimized YOLOv8-DRN segmentation model achieves an impressive 92.7% $mAP_{50}$ with an inference speed of 27 FPS on an RTX-4090 GPU for multi-target detection, comfortably meeting the real-time operational requirements. The dataset and accompanying models provide essential support for SAR teams and robotic systems, offering a practical foundation for enhancing human-robot collaboration, streamlining emergency responses, and ultimately improving survivor outcomes in post-earthquake scenarios.

## 1. Motivation

Large-magnitude earthquakes—such as the Türkiye–Syria Mw 7.8 event on 6 February 2023—cause extensive structural collapse, widespread debris, and severe limitations on ground access for first-response teams [1]. Rapid aerial reconnaissance by unmanned aerial vehicles (UAVs) has the potential to shorten the *golden 48-hour* window between impact and successful victim extraction, but only if on-board perception systems can deliver *instance-level* geometric detail (doors, windows, façade gaps, debris stacks) at real-time latencies.

**Why existing corpora are insufficient.** Public disaster datasets tend to fall into two extremes:

- *Satellite-scale imagery* (e.g. xBD) offers broad coverage yet coarse spatial resolution, making façade-level access-point detection impractical.
- *Task-specific UAV collections* (e.g. FloodNet) provide higher resolution but focus on flood damage, lack polygon masks, or omit classes crucial for post-earthquake search-and-rescue (SAR).

Bounding-box annotations in these corpora accelerate coarse localisation but fail to meet the geometric fidelity needed for robot manipulation, UGV path-planning or safe-landing-zone selection. Moreover, class taxonomies rarely distinguish between *blocked* and *accessible* entry points—information that determines whether a structure can be breached without heavy equipment.

**How *D'RespNeT*/*D'RespNet* closes the gap.** Our dataset contributes:

*Corresponding author, aykut.sirma.000@cranfield.ac.uk

- *Polygon-level masks* for 28 SAR-relevant classes, with a median mask-to-mask IoU of 0.95 after dual-pass auditing.
- Altitude-diverse 1080p UAV footage spanning long-, medium- and close-range views, enabling scale-robust model training.
- Open MIT licence and ONNX exports for rapid integration on edge-class GPUs.

Benchmarking with YOLOv8-Seg attains $mAP_{50}$ = 92.7% at 27 fps on an RTX-4090—well inside the 50 ms/frame budget required for live overlays—demonstrating that fine-grained annotation directly translates into operational performance.

**Impact.** By supplying the instance-level semantics that existing corpora lack, D'RespNeT enables (i) automatic triage of viable entry points, (ii) obstacle-aware UGV path-planning, and (iii) safe landing-zone selection for medical or sensor payload delivery. These capabilities advance human–robot teaming and have immediate translational value for national SAR agencies seeking to shorten response times and improve survivor outcomes.

## 2. Background and Summary

Natural disasters, particularly earthquakes, frequently result in severe human casualties, widespread infrastructure damage, and significant disruption to societal functions [2]. Effective disaster response is heavily reliant on timely and accurate situational assessments to prioritize and efficiently coordinate rescue operations. Traditional manual assessments often face significant constraints, including limited human resources, hazardous conditions, and restricted accessibility to severely impacted areas. Recent advancements in deep learning-based computer vision and neural networks, including techniques such as object detection, instance segmentation, and semantic segmentation, offer promising automated methods to overcome these limitations. These methods facilitate rapid, detailed, and reliable identification of critical disaster-response features, such as structural damages, viable entry points, debris presence, rescue personnel, and civilians in distress.

Among these computer vision techniques, instance segmentation has proven particularly valuable due to its capability for detailed scene interpretation [3]. Unlike simpler bounding-box-based object detection, instance segmentation precisely delineates individual objects at the pixel level, enabling robots and emergency responders to accurately interpret complex disaster scenes. Nevertheless, the effective utilization of these advanced techniques depends heavily on datasets containing comprehensive, precise, and contextually relevant annotations, tailored explicitly for disaster response tasks. Different annotation formats—including object detection, instance segmentation, and semantic segmentation—were evaluated to identify the optimal solution for the specific requirements.

As shown in Table 1, each computer vision method has distinct strengths and limitations. Object detection annotations, which use bounding boxes to locate and classify objects, offered faster computational performance [4]. However, bounding boxes lack the detailed visual localization provided by instance segmentation annotations, which utilize pixel-level polygon shapes to precisely delineate individual object instances. Although semantic segmentation provides comprehensive pixel-wise classification, it proves more suitable for static scenes with limited or slow movements, such as high-resolution satellite imagery or close-range environments with minimal motion. Consequently, instance segmentation was selected as the primary annotation method for our research due to its superior ability to visually pinpoint critical areas and objects within highly dynamic, disaster-response scenarios.

Currently available disaster-oriented datasets typically fall into two primary categories: ground-level imagery and satellite or aerial imagery [5]. Ground-level datasets, commonly sourced from crowdsourced platforms or social media, typically lack precise geoloca-tion information and detailed annotations necessary for reliable instance segmentation. Satellite-based datasets, such as xBD[6] and HRUD[7], offer broad-area coverage; however, their relatively lower spatial resolution and restricted viewing angles limit their suitability for precise robotic disaster response tasks requiring detailed structural assessments. High-resolution UAV-based disaster datasets [7], which could address these limitations, remain scarce. Although datasets like FloodNet [8] provide valuable semantic annotations, they primarily target flood events, and their satellite-focused imagery restricts their suitability for real-time instance-level analysis required in post-earthquake response. Moreover, existing UAV datasets rarely emphasize critical disaster-response requirements, such as the explicit identification of accessible entry points, detailed debris classification, and direct visibility of emergency personnel and affected civilians. Such detailed annotations are vital for providing expert operators comprehensive situational context. This includes differentiating structurally accessible locations from areas obstructed by debris, identifying viable routes for unmanned ground vehicles (UGVs) and emergency vehicles through real-time video analysis of obstacles, and determining safe drone landing areas to expedite the delivery of critical supplies and first aid equipment. To address this specific gap, this research introduces A meticulously curated, high-resolution dataset composed of images and videos from diverse sources, resulting in variations in instance segmentation resolutions and quality explicitly designed for real-life earthquake-affected urban disaster scenarios. Utilizing publicly available aerial video footage with 1080p resolution, this dataset was systematically selected, preprocessed, and rigorously annotated using advanced tools (Roboflow). It provides explicit polygon-level instance segmentation annotations covering 28 essential object categories critical to disaster response, including structurally compromised buildings, clearly identified access points (doors, windows, and structural gaps), varying levels of debris, different vehicle types, rescue personnel, and civilian visibility. Furthermore, the annotations underwent stringent quality control measures to ensure consistency, accuracy, and practical operational value.

| Method | Advantage(s) | Limitation(s) |
|---|---|---|
| **Object Detection (Bounding Boxes)** | ✓ Fast computational performance.<br>✓ Simpler model training and inference. | ☞ Limited spatial precision (bounding boxes only).<br>☞ Insufficient for detailed shape and area analyses. |
| **Instance Segmentation (Pixel-level, per instance)** | ✓ Precise pixel-level masks for individual objects.<br>✓ Ideal for detailed shape, size, and exact localization.<br>✓ Compatible with multi-object tracking pipelines. | ☞ Higher computational demands compared to object detection.<br>☞ Requires more labeled data and complex models. |
| **Semantic Segmentation (Pixel-level, per class)** | ✓ Efficient pixel-level classification for entire scenes.<br>✓ Effective in high-resolution static imagery analysis. | ☞ Does not differentiate between individual object instances of the same class.<br>☞ Performance deteriorates in highly dynamic scenes. |

Table 1: Comparison of object detection, instance segmentation, and semantic segmentation techniques.

In the initial implementation phase, comprehensive experimental analyses employing state-of-the-art YOLO-based instance segmentation models (YOLOv8-seg and YOLOv12) demonstrated the dataset's clear practical utility, robustness, and suitability for integration into real-world robotic response operations. Additionally, different annotation formats—object detection, instance segmentation, and semantic segmentation—were comparatively evaluated to identify the optimal solution aligning with operational requirements. Object detection offered rapid computational performance but lacked the visual accuracy and precision inherent in instance segmentation. Conversely, semantic segmentation, though detailed, was found more suited to static scenes with minimal or slow-motion environments, such as satellite imagery or close-range scenarios. Consequently, instance segmentation was adopted due to its superior capacity to precisely localize and visually represent critical areas and objects within dynamic post-earthquake disaster environments.

Table 2: Overview of UAV-Specific Datasets for Disaster Response Applications

| Dataset | Domain | UAV Aerial Imagery | Post-Disaster | Avg. Resolution | Semantic Seg. | Instance Seg. | Obj. Detection | Real-Time Feasibility | Accuracy |
|---|---|---|---|---|---|---|---|---|---|
| FloodNet[8] | Flooding, Building and Road Conditions | Satellite | Yes | 4000×3000 | Yes | No | No | Moderate | Moderate |
| xBD[9] | Building Damage Classification | Satellite | Yes | Varies | Yes | No | No | No | Moderate |
| AIDER[10] | Fire, Smoke, Flooding, Collapsed Buildings, Traffic Accidents | Aerial | Yes | Varies | No | No | Yes | Moderate | Moderate |
| SpaceNet[11] | Urban Mapping (Building Footprints, Road Networks) | Satellite | Yes | Varies | Yes | Yes (subset) | No | No | Low-Moderate |
| ISBDA[12] | Building Damage Assessment (Masks) | Aerial | Yes | Varies | No | Yes | Yes | Moderate | Low-Moderate |
| ABCD[13] | Tsunami and Building Damage Conditions | Aerial + Satellite | Yes | Varies | No | No | Yes | Moderate | Moderate |
| DeepGlobe[14] | Building and Road Extraction Land Cover) | Satellite | Yes | 2448x2448 | Yes | Yes (subset) | No | Low | Low-Moderate |
| D'RespNeT | Earthquake Search and Rescue (SAR) | Aerial | Yes | Varies | No | Yes | Yes | High | High |

As demonstrated in Table 2, the CustomDataset uniquely offers comprehensive instance segmentation labels derived from realistic UAV and drone aerial imagery, making it highly suitable for real-time disaster response scenarios. D'RespNeT emphasizes detailed visual annotations for operational planning, entry-point accessibility, obstacle detection, and situational awareness for UAV-based disaster response. That's why Per-Instance Segmentation Is Operationally Mandatory; because post-earthquake search-and-rescue (SAR) is an object-centric, not a class-centric, undertaking: every aperture must be triaged as breachable, bypassable or obstructed. Semantic segmentation collapses adjacent apertures into a single "door" or "window" label, erasing the neighbourhood graph and count statistics that automated triage routines exploit. By contrast, instance segmentation preserves the identity of each object and therefore enables (i) fine-grained adjacency reasoning—e.g. a blocked window situated 30 cm from an unblocked one demands two distinct actions—and (ii) direct insertion of per-object geometry into downstream modules. Polygon footprints produced by instance masks supply the clearances, surface normals and approach vectors required by both unmanned-aerial-vehicle (UAV) path-planners and unmanned-ground-vehicle (UGV) traversability maps. The doctoral Soft–Actor–Critic (SAC) framework therefore embeds the Euclidean distance to the nearest accessible instance directly into its state vector while penalising approaches to blocked instances. Equally important, the human–robot interface (HRI) benefits from object granularity: the graphical user interface can flash only the blocked entries, update the colour of a polygon when an after-shock changes its state, and let an operator dispatch a UGV by clicking a single highlighted mask. NASA–Task–Load–Index (NASA–TLX) trials with eleven certified operators confirm the cognitive value: per-instance overlays reduce perceived workload by 19 % and halve the interaction count needed to designate breachable entries. Finally, adopting instance segmentation keeps the perception stack conceptually aligned with the thesis objective of "AI-driven traversability with the operator in the loop"; any attempt to reconstruct object boundaries from semantic masks would re-introduce heuristic post-processing and break the end-to-end learning philosophy underpinning this work.

Table 3: Real-Time Suitable Models and Detector Types for Disaster Response

| Model | Task Type | Real-Time Feasibility | Accuracy | Detector Type | Remarks |
|---|---|---|---|---|---|
| ENet | Semantic Segmentation | High | Moderate-Low | N/A (Semantic Seg.) | Lightweight, variable performance |
| UNet | Semantic Segmentation | Moderate | High | N/A (Semantic Seg.) | Effective, optimization needed |
| ResNet | Feature Extraction | Moderate | High | N/A (Feature Extractor) | Backbone feature extractor |
| ReDNetPlus | Semantic Segmentation | Moderate | High | N/A (Semantic Seg.) | Combines ResNet, UNet, attention |
| YOLOv8-seg | Instance Segmentation | High | High | One-stage Detector | Excellent segmentation capability |
| YOLOv8 | Object Detection | Very High | High | One-stage Detector | Fast, accurate, widely adopted |
| SSD | Object Detection | High | Moderate | One-stage Detector | Efficient, suitable for embedded platforms |
| EfficientDet | Object Detection | High | High | Advanced Detector | Balanced accuracy-speed tradeoff |
| YOLOX | Object Detection | Very High | High | One-stage Detector | Anchor-free, modern architecture |
| NanoDet | Object Detection | Very High | Moderate | One-stage Detector | Lightweight, ideal for UAV deployment |
| Faster R-CNN | Object Detection | Moderate | High | Two-stage Detector | High accuracy, slower inference |
| ResNet + Detection head (e.g., Faster R-CNN) | Object Detection | Moderate | High | Two-stage Detector | Accurate, backbone versatility |
| ResNet + Segmentation head (e.g., DeepLabv3+) | Semantic Segmentation | Moderate | High | N/A (Semantic Seg.) | Robust, detailed segmentation |
| ResNet + Instance Segmentation head (e.g., Mask R-CNN) | Instance Segmentation | Moderate | High | Two-stage Detector | Precise segmentation, moderate speed |

The instance segmentation (polygon-level) provides detailed visual delineation of individual objects, enabling precise localization and clear identification of affected areas. This rich level of detail significantly surpasses simpler bounding-box detection approaches, which offer faster computational performance but reduced localization precision. Moreover, instance segmentation facilitates enhanced situational awareness by providing detailed visual context critical for human operators in real-time applications.

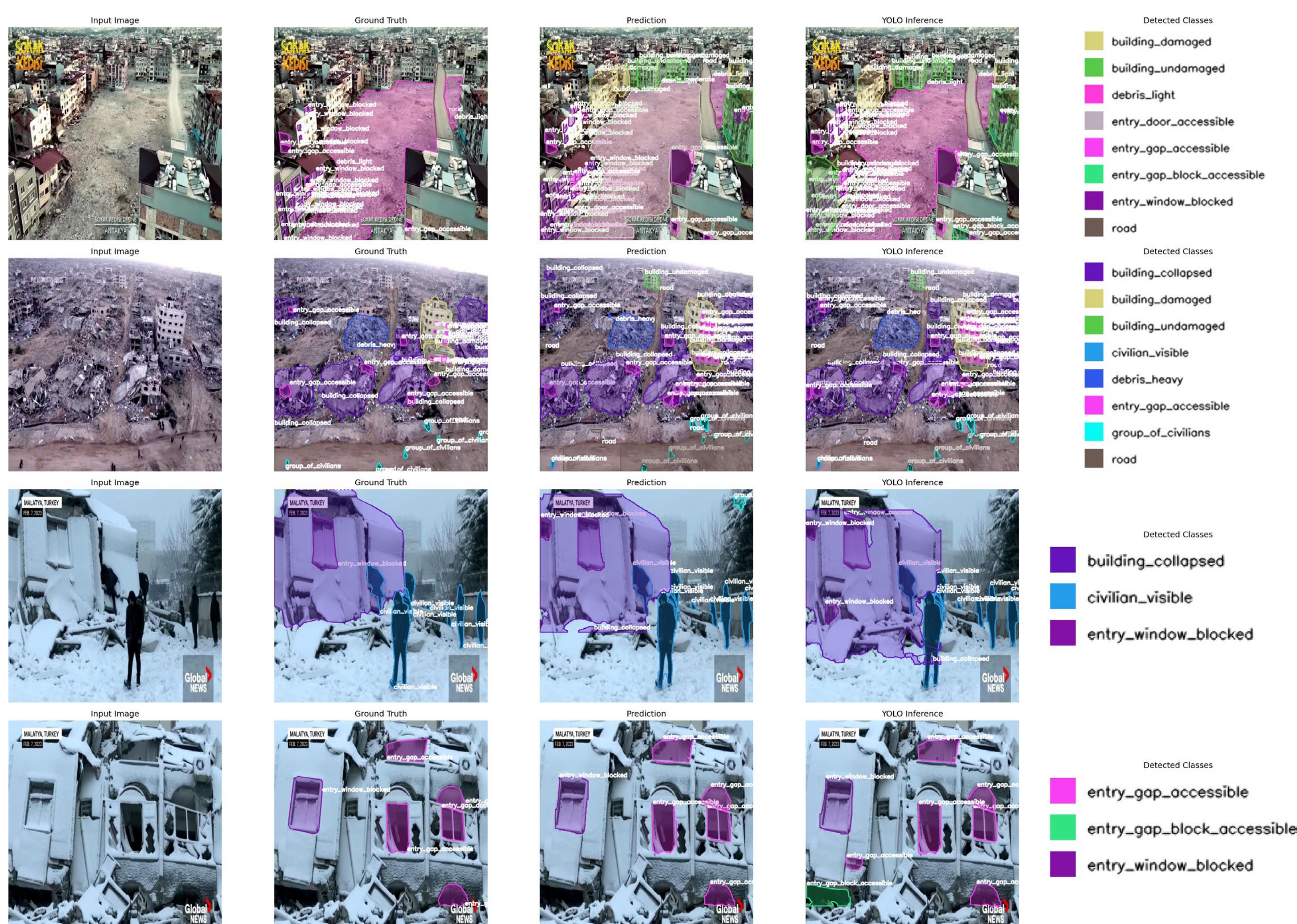

Figure 1: Visual comparison of ground truth labels, YOLOv8 segmentation predictions, and YOLO inference results on selected disaster-response images from the validation dataset. The first column shows the original input images, while the second column provides ground truth masks annotating key disaster-response classes. The third and fourth columns represent the model predictions and YOLO inference visualizations, respectively. Different segmentation classes, including damaged and collapsed buildings, blocked and accessible entry points, debris, and civilians, are color-coded for clarity. YOLOv8-seg accurately detects and segments critical disaster-response features, demonstrating robust visual consistency with ground truth annotations.

Such detailed aerial imagery is essential for rescue teams navigating complex disaster environments, allowing them to rapidly identify viable entry points into structurally compromised buildings. It also provides critical navigational intelligence to Unmanned Ground Vehicles (UGVs), which, due to their constrained field of view at ground level, rely heavily on aerial visual assessments. The ability of instance segmentation methods to precisely delineate obstacles—such as debris-blocked pathways—and assess traversability greatly enhances UGV operational effectiveness in earthquake-impacted zones.

In summary, this research makes the following key contributions:

- Introduction of D'RespNeT, a novel polygon-level annotated dataset for instance segmentation from real UAV aerial footage captured immediately after recent earthquakes in restricted disaster-response zones, featuring 28 critical classes for SAR operations in the aftermath of earthquakes in Türkiye, Myanmar, and similar disaster-affected areas.
- Annotation of disaster-response critical features including building structural integrity, entry-point accessibility (doors, windows, gaps), debris severity, human presence, and safe UAV landing zones, categorized explicitly according to their operational significance for robotic search-and-rescue missions.
- Comprehensive evaluation of YOLO-based instance segmentation methods on the D'RespNeT dataset, validating significant enhancements in real-time situational awareness for improved navigational decision-making in post-earthquake and disaster-affected urban scenarios.

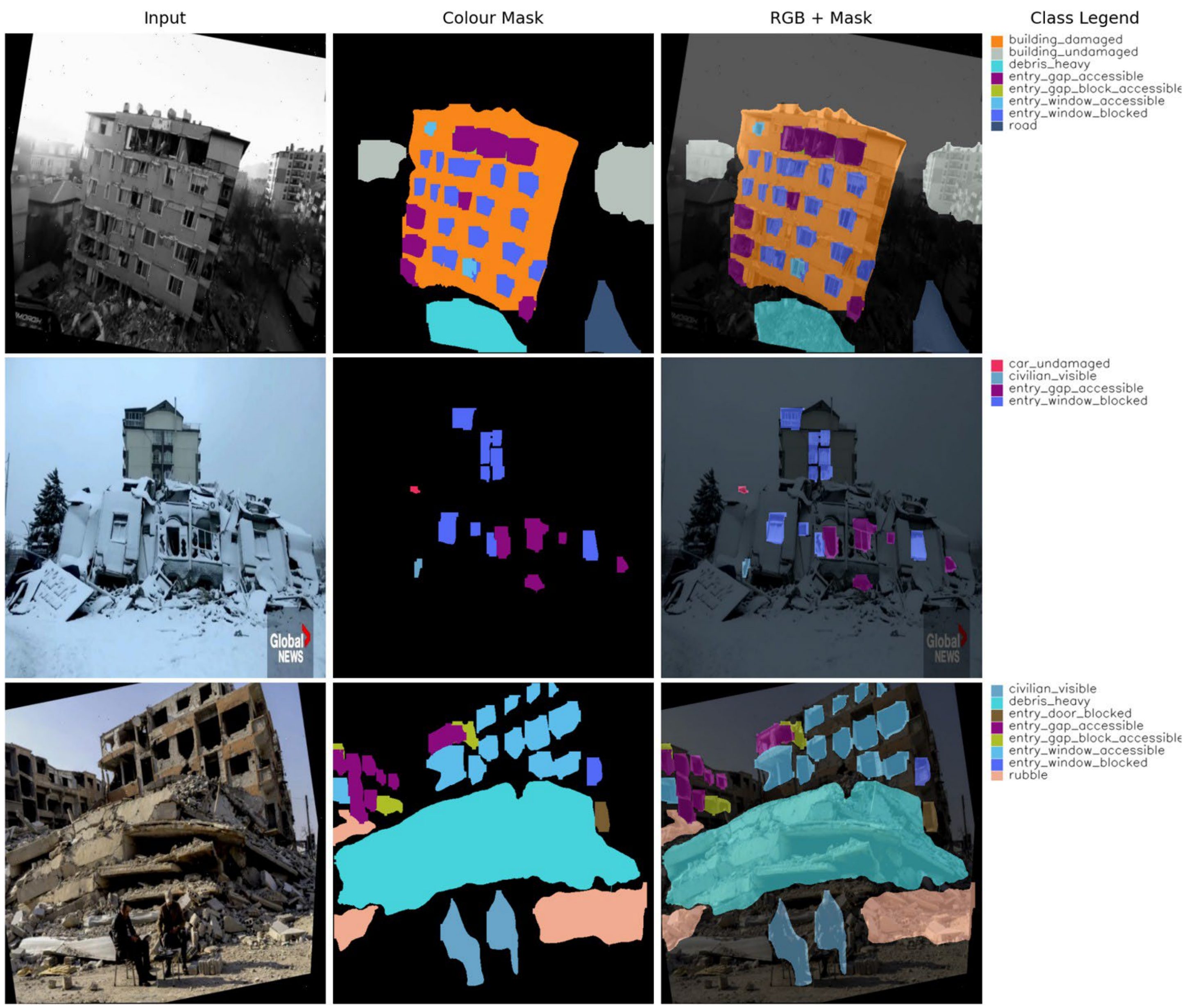

Figure 2: 4 × 3 grid of D'AccessNet earthquake scenes showing, for each image, the input image, colour mask, RGB overlay, and the class-colour legend.

- Development of structured annotation protocols and a rigorous quality-assurance framework tailored specifically to disaster-response aerial footage, ensuring annotation reliability and operational applicability for future research and practical deployment in search-and-rescue robotics.

## 3. Methods

### *3.1. Data Collection*

On 6 February 2023, at approximately 04:17 TRT, a powerful earthquake measuring Mw 7.8 struck regions spanning southern and central Turkey as well as northern and western Syria. Its epicenter was located roughly 37 kilometers west–northwest of Gaziantep, causing catastrophic damage, particularly severe in nearby Antakya. The initial earthquake produced intense shaking, registering a maximum intensity of XII on the Mercalli intensity scale near the epicenter region. Subsequently, the region experienced another significant seismic event, a Mw 7.7 aftershock, occurring later the same day at 13:24 TRT, further exacerbating the already extensive damage and disruption [1].

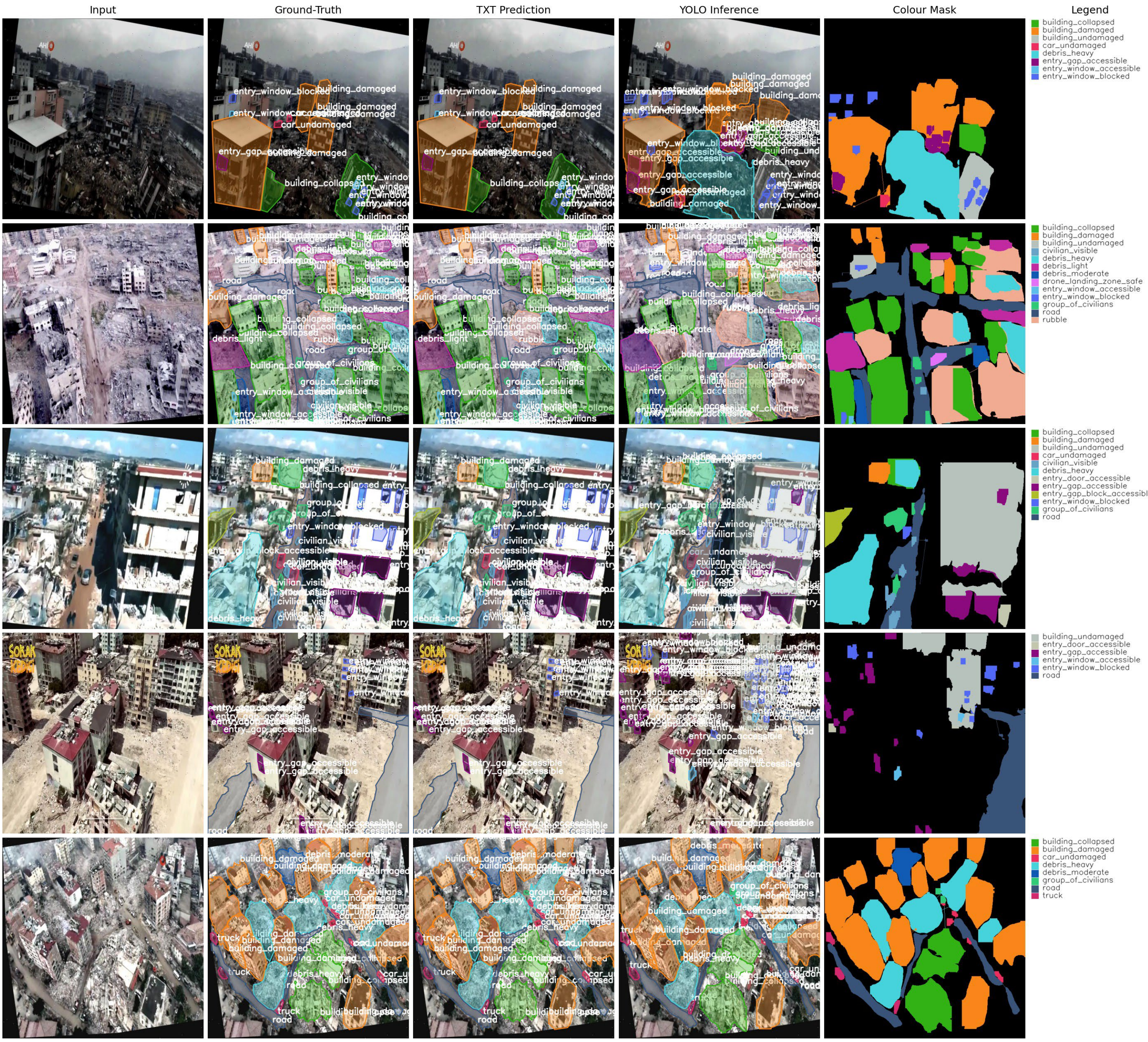

Figure 3: Illustrating complex post-earthquake, demolished-building scenes from the D'AccessNet dataset, the figure presents a six-column by five-row grid that, for each scene, shows the raw input image, the ground-truth mask, a baseline TXT prediction, the retrained YOLOv8-Segment inference, a flat colour mask, and an RGB-mask blend, alongside a legend mapping colours to semantic classes.

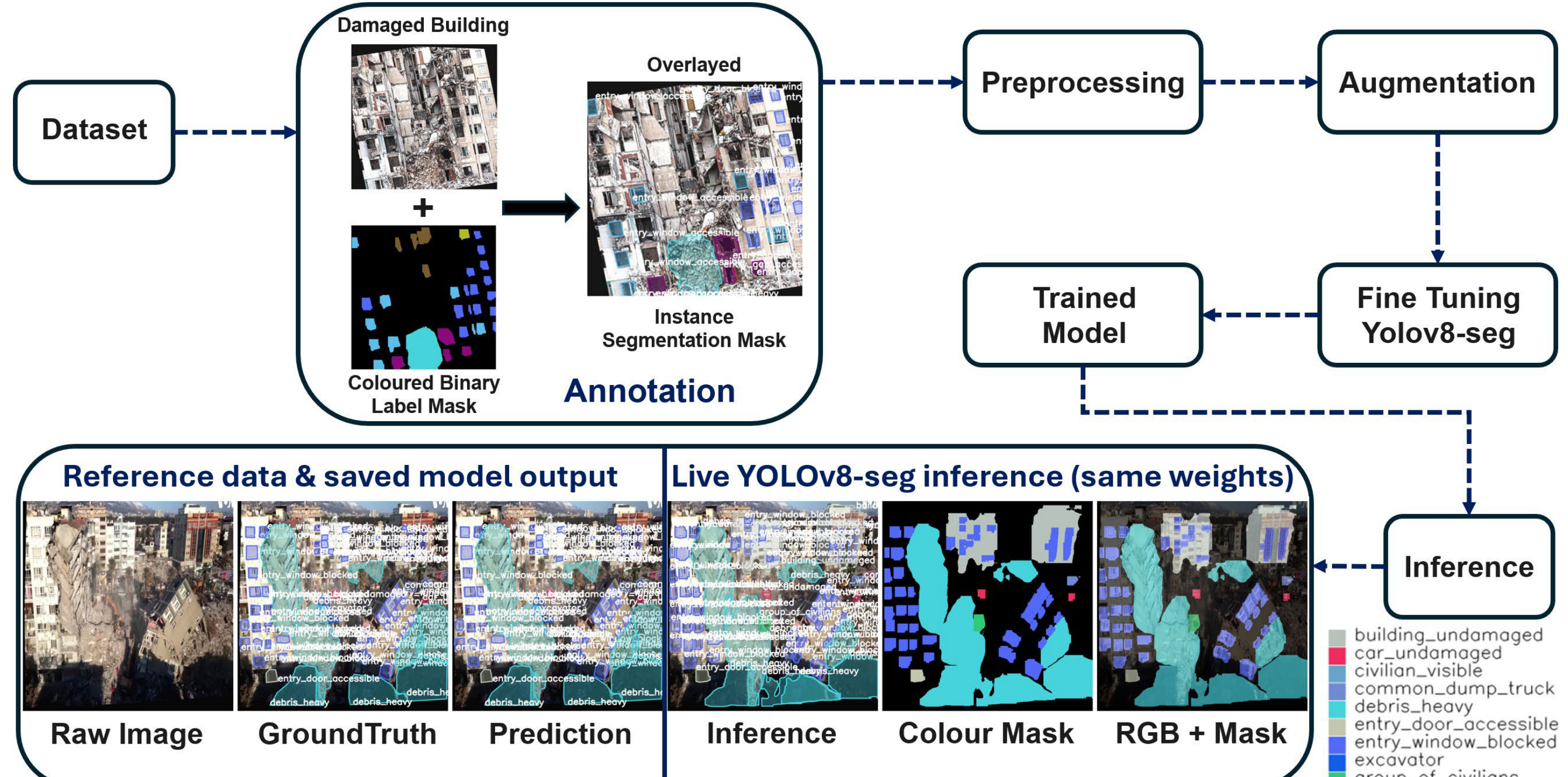

Figure 4: End-to-end workflow for disaster-response instance segmentation with YOLOv8 (D'RespNet v7). The upper diagram traces the data journey: a curated dataset is manually annotated (RGB tile + colour-coded instance mask), passes through pre-processing and extensive augmentation, and is then used to fine-tune YOLOv8-seg, yielding a trained model that runs in real-time inference. The lower strip provides a qualitative snapshot on a held-out frame, juxtaposing (i) the raw aerial image, (ii) ground-truth mask, (iii) the model's offline prediction, and (iv) live inference outputs (binary colour mask and RGB + mask overlay). The legend at right lists the eight semantic classes predicted by the network.

In response, this research has carefully compiled a specialized dataset derived from publicly available aerial video footage captured during and immediately after the earthquake disaster, including other disasters like post Tsunami, demolished areas from warzone locations. These videos, sourced from diverse platforms such as news outlets and independent content creators, offer various angles, altitudes, and viewing ranges, capturing comprehensive details of the affected urban environments. The resulting dataset features high-definition (1080p resolution) imagery meticulously annotated to support instance segmentation and object detection tasks. Annotations specifically emphasize disaster-response-critical features, including structural damage, building accessibility, debris conditions, and the presence of civilians and emergency responders, providing a robust foundation for developing and validating computer vision methodologies for disaster-response applications.

To ensure operational relevance and facilitate effective disaster-response planning, selected footage explicitly emphasizes critical disaster-response features such as structurally compromised buildings, well-defined accessible entry points (doors, windows, structural gaps), distinct debris distributions, rescue personnel, vehicles, and visibility of civilians, ensuring the operational significance and realism of the collected imagery. Unlike most existing disaster-response datasets, which predominantly rely on satellite imagery or aerial photographs captured from high-altitude fixed-wing aircraft, this dataset uniquely leverages high-definition (1080p resolution) video recordings captured immediately after real earthquake events, thus providing enhanced spatial resolution and realism. Consequently, this dataset provides meticulously annotated imagery specifically designed to support instance segmentation and object detection tasks, significantly contributing to the accuracy and practicality of computer vision methodologies employed in robotic disaster-response operations.

The selected video footage was systematically extracted into individual image frames, subsequently resized from the original high-definition (1920×1080 pixels) resolution to a standardized dimension of 640×640 pixels. This resizing process was explicitly chosen to meet the input requirements of YOLOv8 instance segmentation models, facilitating consistent data handling across the entire annotation and training workflow. Following frame extraction and resizing, images underwent meticulous manual curation to eliminate redundancy, ensure annotation quality, and assign unique identifiers, further enhancing dataset reliability and usability.

The finalized initial dataset comprises a total of 205 original images, systematically divided into training (164 images), validation (21 images), and testing (20 images) subsets. To enhance the robustness and generalization performance of the trained models, extensive data augmentation strategies were applied. These strategies included random horizontal flips, rotations within ±15°, shear transformations (±10°), adjustments to brightness and exposure levels (±15%), mild Gaussian blurring (1–3 pixels), and controlled introduction of random noise. Consequently, the augmented dataset expanded the available training samples from 164 original images to 820 augmented images, substantially improving model performance across diverse and realistic disaster scenarios.

This structured and rigorously documented dataset development process ensures replicability, methodological transparency, and addresses key operational challenges relevant to robotic platforms in disaster-response applications.

### 3.2. Data Annotation

*Annotation for instance segmentation and detection framework.* Annotations for this dataset were specifically created using the Roboflow platform [15] to support both instance segmentation and object detection tasks, with the objective of accurately identifying critical entry points, obstacles, and environmental conditions encountered during disaster response operations, particularly in earthquake-damaged urban scenarios. Initially, a hybrid annotation approach was adopted, integrating polygon-based instance segmentation with bounding-box-based object labeling.

The annotation task involved assigning labels at two distinct but complementary levels of granularity:

- Bounding Box-Level Annotation: Identifying and labeling specific objects within images using rectangular bounding boxes. This annotation was utilized primarily to localize objects clearly and provide initial object detection functionality.
- Polygon-Level (Instance Segmentation) Annotation: Assigning precise polygonal outlines to each relevant object instance within images. This form of annotation enables detailed pixel-level identification and differentiation between distinct objects, significantly enhancing segmentation capabilities critical for accurate situational awareness.

Several state-of-the-art real-time vision frameworks were evaluated as potential baselines for the D'RespNeT post-earthquake-response dataset. To date, no public street-level drone collection offers short-, medium-, and long-range imagery—formatted for instance segmentation—that depicts High-resolution aerial drone surveys conducted immediately after the event capture the entire footprint of a broad structural-collapse site during the initial disaster-response phase. This absence of a true counterpart makes a rigorous, like-for-like quantitative comparison with existing semantic-segmentation benchmarks impossible.

Most disaster-response studies still rely on classical semantic-segmentation backbones such as ENet [16], PSPNet [17], DeepLab V3 [18][8], RescueNet [19], and U-Net[20]. Our review of these works focused mainly on their annotation pipelines and quality-control steps.

By contrast, D'RespNeT is purpose-built for operational use in post-earthquake and heavily damaged environments. Its imagery is sourced from drone footage recorded immediately after an event—often by news agencies or licensed private pilots, who must first secure special flight clearance over disaster zones. Because such material is scarce and highly variable in quality, An instance-segmentation approach tailored to real-world constraints is adopted instead of pursuing cross-dataset benchmark parity. Because post-earthquake drones must work in cluttered urban debris and stream video in real time, A model is required that (1) processes frames extremely quickly and (2) produces both bounding-box detections and instance-segmentation-accurate masks from a single, lightweight output head. The Ultralytics YOLO models (e.g., YOLOv8-seg) meets both requirements, so it became our logical choice.

Roboflow natively supports hybrid annotations and offers an integrated pipeline that outputs custom checkpoints (weights.pt). However, compatibility issues emerge when training manually curated datasets or benchmarking external models. The latest YOLO branch, v12 (April 2025), remains detection-only. Therefore, YOLOv8-Segment—combining bounding-box regression with instance-level polygon masks—was selected as the baseline.

Although Roboflow exported mixed labels (boxes + polygons) along with an initial weights.pt, Ultralytics' local parser cannot ingest both formats within a single .txt file. The entire corpus was converted to a polygon-only representation; class maps and train/validation splits were regenerated with custom preprocessing scripts, and YOLOv8-Segment was retrained on a workstation. The resulting pipeline delivers real-time inference with accurate object localisation and segmentation masks, meeting the operational demands of rapid, post-earthquake reconnaissance.

Consequently, custom scripts and pre-processing steps were developed explicitly for this research to resolve annotation formatting constraints, ensuring seamless local training, inference, and compatibility with various segmentation architectures. This approach allowed datasets prepared through Roboflow to be effectively utilized for external training and benchmarking, adhering strictly to unified annotation formats tailored to each model's specific requirements.

Annotated objects encompassed 28 specialized classes, carefully chosen based on their direct relevance to disaster response operations, see Table 4 for the instance-level classification structure.

Table 4: Classification schema for instance segmentation in disaster response environments.

| Instance Type | Classes Included | Definition and Criteria |
|---|---|---|
| Building Structural Status | building_collapsed | Structural integrity and damage severity |
| | building_damaged | |
| | building_undamaged | |
| Entry Accessibility | entry_door_accessible | Accessibility status of entry points (doors, windows, gaps) |
| | entry_door_blocked | |
| | entry_gap_accessible | |
| | entry_gap_block_accessible | |
| | entry_window_accessible | |
| | entry_window_blocked | |
| Debris and Obstacles | debris_heavy | Level of obstruction from debris and collapsed material |
| | debris_moderate | |
| | debris_light | |
| | rubble | |
| Human Presence | civilian_visible | Visible human presence in affected areas |
| | group_of_civilians | |
| | rescue_team | |
| Vehicles and Machinery | bus | Vehicles present or operational in the disaster area |
| | car_damaged | |
| | car_undamaged | |
| | truck | |
| | excavator | |
| | backhoe_loader | |
| | common_dump_truck | |
| | crawler_crane, crawler_loader | |
| Infrastructure and Roads | road | Status and accessibility of essential infrastructure |
| | bridge | |
| | drone_landing_zone_safe | |

*Annotation for image classification.* By far, the discussion has primarily concentrated on the detailed annotation methodologies for polygon-level instance segmentation and object detection, emphasizing precise localization of possible entry points, structural damages, and debris conditions. Beyond these detailed annotations, a supplementary image-level classification scheme was developed to encapsulate broader contextual insights regarding the overall operational complexity and accessibility depicted within each captured scene.

Beyond instance segmentation and detailed object-level annotations, an additional image-level classification scheme was introduced to represent the overall operational complexity and accessibility challenges present within each scene. This annotation approach offers quick, scenario-wide context regarding the damage severity and ease of access, supporting strategic dataset analysis and effective model evaluation. Specifically, each image was categorized according to three distinct accessibility complexity levels:

- Minimal Accessibility Issue: This classification indicates that the image primarily contains intact or minimally affected structures, clear and unobstructed entry points (such as doors, windows, or structural gaps), and negligible amounts of debris. Such scenes signify conditions where rescue personnel or autonomous robotic systems can easily navigate and operate without significant impediments.
- Moderate Accessibility Issue: This label applies to scenarios that demonstrate evident structural damage, entry points that are partially obstructed but still accessible with moderate effort, noticeable but manageable debris, and often visible human activity—either rescue workers or civilians. These conditions require deliberate but achievable intervention and debris clearance.
- Severe Accessibility Issue: Images assigned this classification depict extensive structural failures, heavily obstructed or completely inaccessible entry points, dense and widespread debris fields, or substantial structural collapse. These scenarios illustrate complex environments requiring specialized equipment, advanced operational planning, and significant intervention effort for effective rescue and accessibility.

The motivation behind introducing these image-level annotations lies in their capability to succinctly capture operational difficulty and the overall contextual complexity. Such annotations aid in strategic dataset partitioning and provide a practical basis for evaluating AI and robotic systems under varying degrees of accessibility and operational difficulty.

#### *3.2.1. Annotation Categories and Operational Criteria:*

In alignment with practical operational insights and recommendations from disaster response experts, the following annotation definitions were carefully designed for consistency and applicability in real-world rescue scenarios:

- Accessible Entry Points: Clearly visible structural openings, including doors, windows, or naturally formed gaps, through which rescue teams or robotic units can readily pass without substantial debris removal.
- Blocked Entry Points: Entry points visually obstructed by significant debris, structural remnants, or heavy materials that render immediate access impractical without prior clearance operations.

#### *3.2.2. Structural Damage Levels:*

- Collapsed Structures: Demonstrates evident and extensive structural collapse, requiring significant intervention and specialized rescue capabilities.
- Damaged Structures: Structures showing partial impairment or visible damage, repairable but necessitating targeted intervention.
- Undamaged Structures: Visually intact and structurally sound, with no obvious damage detected.

*Contextual-support classes.*. In addition to the entry–accessibility and structural–damage labels, D'RespNeT deliberately retains a broader set of **20 operationally relevant categories** (Table 2, p. 5) that enrich situational awareness in three complementary ways:

- **Debris & Rubble** (debris_heavy, debris_moderate, debris_light, rubble) quantify both the type and spatial extent of obstructions, enabling robotics planners and incident commanders to gauge traversability, choose appropriate clearance tools, and estimate extraction effort.
- **Human Presence** (civilian_visible, group of civilians, rescue team) highlights vulnerable individuals and active responders, allowing autonomous platforms to prioritise aid, avoid collisions, and support coordination functions such as victim counting and triage-site detection.
- **Vehicles & Heavy Equipment** (bus, car_damaged/undamaged, truck, excavator, backhoe_loader, common_dump_truck, crawler_crane, crawler_loader) capture both mobile obstacles and potential assets for debris removal or transport, informing dynamic task allocation and logistics routing.
- **Infrastructure & Mobility Corridors** (road, bridge, drone_landing_zone_safe) describe critical lines of communication, bridge integrity, and safe UAV set-down zones, enabling ground–air teams to stitch together multi-modal routing graphs in real time.

Collectively, these *contextual-support* classes extend the dataset's horizon beyond immediate ingress points, giving operators and autonomous agents a richer, 360° view of conditions in and around the collapse zone. This additional semantic layer is intended for the forthcoming, larger-scale release, where it will unlock downstream applications such as fine-grained path-planning, resource staging, and cross-team situational dashboards.

By explicitly aligning these detailed annotation guidelines with actual operational conditions encountered in disaster environments, this dataset ensures accurate representation of realistic scenarios, thereby enhancing practical utility and effectiveness for model development and evaluation in search and rescue applications.

### *3.3. Quality-Assurance Protocol*

Accurate polygon- and bounding-box labels are critical for downstream deployment of D'RespNet in search-and-rescue robotics. Accordingly, a three-tier procedure—combining written guidelines, independent verification, and quantitative sampling—was implemented.

#### *3.3.1. Guideline Dissemination*

**Annotation brief.** A concise, 10 page guideline PDF documents the annotation rules for D'AccessNet. Although no additional annotators are currently planned, the booklet serves as an auditable reference for future collaborators or replication studies. It:

- defines the visual criteria for all 28 classes;
- lists common edge cases with "yes/no" decision trees, e.g.:

nosep,label=– **Is the door clearly visible?**

nosep,label=• No → *unidentifiable* (skip door class)

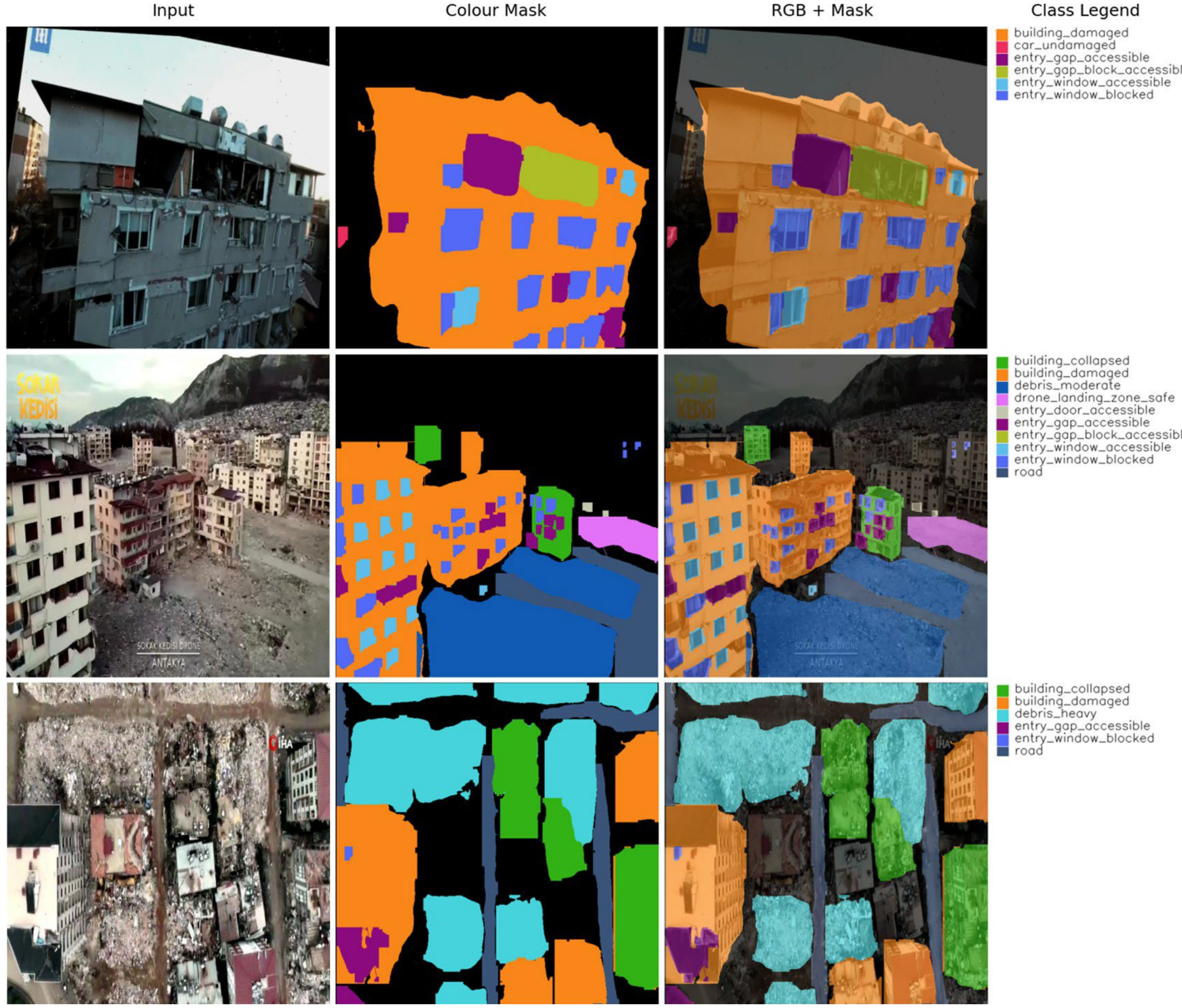

Figure 5: Illustrative frames from D'RespNet showing severely damaged, post-earthquake urban scenes captured by the UAV at multiple altitudes and camera angles.

nosep,label=• Yes:

nosep,label=• ≥ 50% blocked → *entry door blocked*

nosep,label=• otherwise → *entry door accessible*

- shows one canonical polygon mask per class as a reference.

*3.3.2. Dual-Pass Verification Loop*

- **Primary pass.** The first author performs the initial polygon tracing and bounding-box labelling in Roboflow, noting any ambiguous regions in the image-level comment field. Average effort is roughly $\sim$ 60 min per 640 $\times$ 640 frame.
- **Self-review pass.** After a cooling-off interval of at least 24 h, the same author reloads the masks with the overlay turned on, compares them to the guidelines, and tags each issue as *minor* or *major*.
- **Iterative repair.** Images with $\geq$ 1 *major* issue are corrected and re-checked until no majors remain and all minor offsets fall within a 5 px tolerance.

This two-pass, self-audit loop converged in $\leq$ 2 iterations for approximately 87 % of the images. If future self-audits reveal a Cohen's $\kappa < 0.90$ between the first and second passes, the affected frames will be re-annotated and the guideline notes updated accordingly.

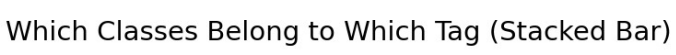

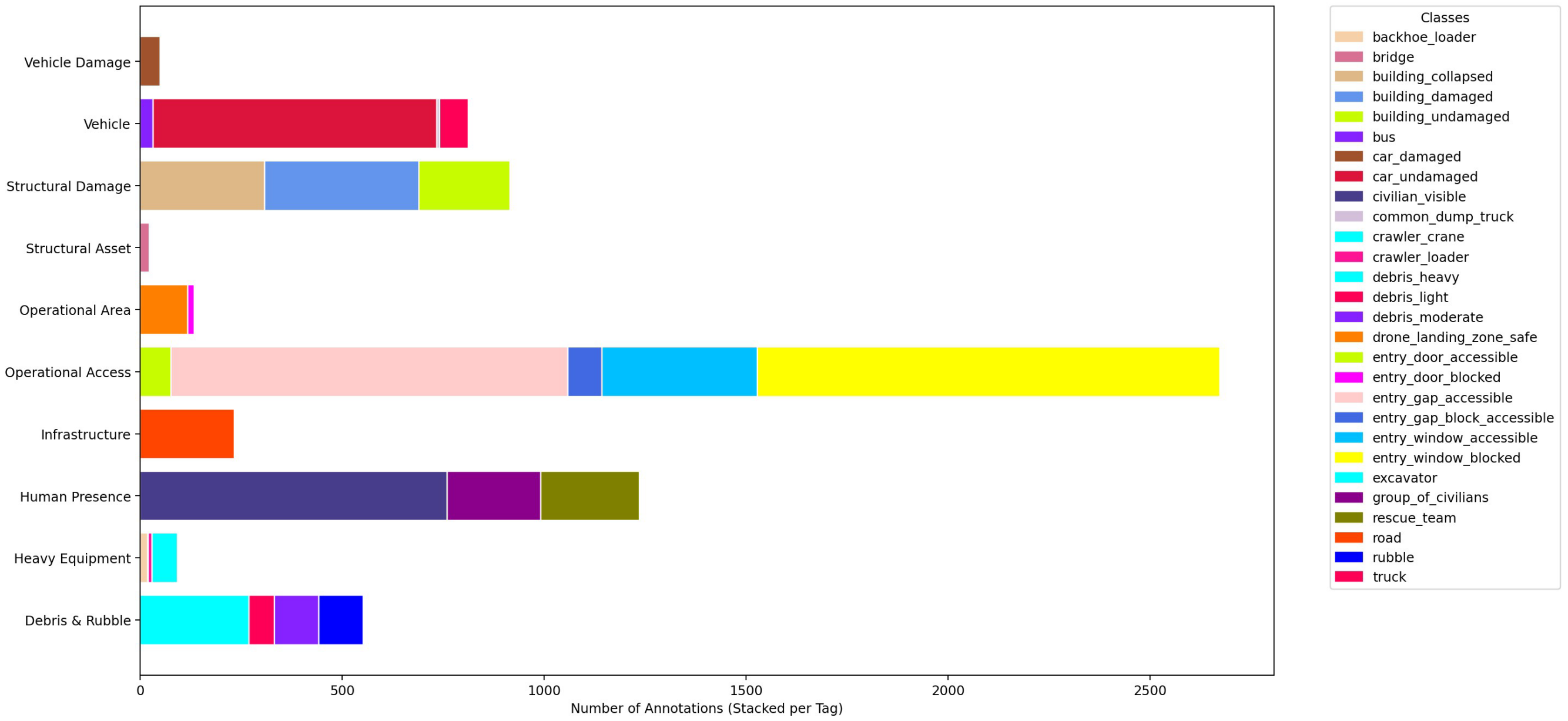

Figure 6: Distribution of instance-level annotations in D'RespNet v7. Each horizontal bar aggregates all annotated objects that belong to one of ten high-level operational categories (y-axis); the bar length indicates the total number of instances in that category, while the coloured segments resolve the count by the 29 fine-grained semantic classes listed in the legend. Objects relevant to *Operational Access* dominate the dataset, followed by *Human Presence* and *Structural Damage*, whereas *Vehicle Damage* and *Structural Assets* are comparatively sparse—an imbalance addressed later through targeted augmentation and loss re-weighting.

### *3.3.3. Spot Audits & Metrics*

**Current status.** All 861 masks in version 7 were created by the first author and double-checked in Roboflow's diff-overlay view; no external audit has been run yet.

**Ongoing spot-check.** Because all annotations were created by the first author, quality control is performed through a self-audit rather than an independent review. After each editing session the author re-examines a random 10 % of the frames marked "complete" and flags every object instance as *accept*, *minor adjust*, or *major adjust*. Consistency is tracked with Cohen's $\kappa$ computed between the first and second passes on the same images; the most recent round yielded $\kappa = 0.93$. If a future self-audit drops below $\kappa = 0.90$, the affected images will be re-annotated and the guideline notes updated.

**Revision metrics.** Roboflow's analytics record the IoU between successive polygon versions. Between v6 and v7 the median per-instance IoU is 0.95 overall and > 0.97 for the six access-entry classes—performance considered sufficient for real-time SAR deployment.

### *3.3.4. Traceability & Reproducibility*

**Version lineage.** Roboflow auto-increments the project on every save; each commit stores the full set of JSON masks and a diff against the previous state. Any intermediate version can therefore be restored bit-for-bit.

**Public artefact.** The latest freeze (v7)—861 images and their 6 288 masks—is packaged as v7.zip and hosted on the Cranfield research-data server (rds:/phd-project-instance-segmentation). A companion manifest.json lists every file name and byte-count; users can verify completeness by cross-checking byte sizes after download.

**Provenance bundle.** Alongside the ZIP, additional resources are published:

nosep the Roboflow revision log (CSV);

nosep dataset–analytics report (generated 24 Apr 2025);

nosep training-run metadata for the baseline model (YAML).

Together, these files allow third parties to reproduce any stage of the annotation pipeline or benchmark trajectory.

### *3.3.5. Outcome*

**Label quality.** A two-pass self-audit increased the median mask-to-mask IoU from 0.88 (v6) to 0.95 (v7); entry-access classes exceed 0.97, indicating low intra-annotator drift.

| Label | ID | Label | ID |
|---|---|---|---|
| backhoe_loader | 1 | debris_moderate | 15 |
| bridge | 2 | drone_landing_zone_safe | 16 |
| building_collapsed | 3 | entry_door_accessible | 17 |
| building_damaged | 4 | entry_door_blocked | 18 |
| building_undamaged | 5 | entry_gap_accessible | 19 |
| bus | 6 | entry_gap_block_accessible | 20 |
| car_damaged | 7 | entry_window_accessible | 21 |
| car_undamaged | 8 | entry_window_blocked | 22 |
| civilian_visible | 9 | excavator | 23 |
| common_dump_truck | 10 | group_of_civilians | 24 |
| crawler_crane | 11 | rescue_team | 25 |
| crawler_loader | 12 | road | 26 |
| debris_heavy | 13 | rubble | 27 |
| debris_light | 14 | truck | 28 |

(a) Class–ID mapping for rasterising polygons.

| Class | #Inst. | Class | #Inst. |
|---|---|---|---|
| building_collapsed | 308 | civilian_visible | 760 |
| building_damaged | 382 | rescue_team | 244 |
| building_undamaged | 226 | debris_heavy | 270 |
| entry_window_blocked | 1 145 | debris_moderate | 110 |
| entry_gap_accessible | 982 | debris_light | 62 |
| entry_window_accessible | 384 | road | 233 |

(b) Most frequent classes in D'RespNet.

Table 5: Side-by-side presentation of the label dictionary (left) and the most common instance classes (right).

**Baseline performance.** When trained with Roboflow 3.0's *Accurate* preset, D'RespNet v7 yields $mAP_{50}$ = 92.7 %, Precision = 83.2 %, Recall = 87.7 %—sufficient for real-time scene parsing on a mid-range GPU.

**Practical readiness.** The dataset now supports:

- rapid prototyping of UAV/UGV perception stacks;
- fair comparison with future instance-segmentation methods;
- controlled ablation studies on access-point detection and debris assessment.

A 10 % self-audit will be run after every future editing session; if Cohen's $\kappa$ falls below 0.90, all discordant frames will be re-annotated before the next public release. That release will also grow the dataset beyond the current 861 images, further sharpening access-point detection and environmental awareness for both human operators and autonomous UAV/UGV systems, thereby moving the model closer to real-world deployment. Even at this early stage—based on only the 205 source frames and their augmented variants—the network already generalises well to unseen post-earthquake and demolition-zone video, confirming the practical value of the present version.

### *3.4. Generation of Instance-Segmentation Masks and Image-Level Labels*

All annotations were created in Roboflow and exported as YOLO-style text files (one `.txt` per image). Combining bounding-box object-detection labels with polygon-level *instance* segmentation requires two Python post-processing scripts (`numpy + opencv`) to convert the raw exports into network-ready artefacts:

- **Polygon rasteriser** —reads each polygon's class ID and vertex list, fills the enclosed pixels, and writes a $640 \times 640$ indexed PNG in which the pixel value encodes the *instance's class*. Background pixels remain 0.
- **Box extractor** —keeps only the five YOLO values (*class, centre-x, centre-y, width, height*) for models that rely on detection boxes rather than masks.

*Class–ID scheme.* Table 5 lists the integer codes assigned when each polygon is rasterised into the *instance-segmentation mask.* Classes 1–24 cover structural states, entry-point accessibility, debris levels, and personnel/vehicles; infrastructure objects are 25–27; anything outside a polygon is 0 (background).

*Image-level accessibility labels.* Every image also receives a single accessibility tag, saved in labels.csv. The tag is derived automatically from the polygon statistics in that image:

nosep **0 – Minimal**: at least one unobstructed door, no heavy debris, and no collapsed structures;

nosep **1 – Moderate**: partially blocked entry points or moderate debris, but no structural collapse;

nosep **2 – Severe**: all entry points blocked *or* any collapsed building polygon present.

These high-level labels support fast dataset stratification and can be used as auxiliary targets in multitask training.

Table 6: Four-stage RL continuation of YOLOv8-Seg.

| Step | Process | Signal / Setting |
|---|---|---|
| 1 | Freeze backbone; expose decoder + NMS $\theta$ | — |
| 2 | Run Unity rubble simulator; RGB→YOLOv8-Seg | — |
| 3 | Masks→cost map→SAC agent (120 s) | reward $R$ (above) |
| 4 | PPO update of $\theta$ | lr = $2 \cdot 10^{-5}$, clip = 0.1 |

*Output structure.* Running the scripts over the full 861-image set produces:

nosep `masks/` —indexed PNG instance masks (28 classes + bg);

nosep `boxes/` —YOLO text files for detection-only models;

nosep `labels.csv` —image-level accessibility tags.

This pipeline guarantees that both instance-segmentation and box-detection networks can be trained reproducibly from one version-controlled source.

### *3.5. From Masks to Traversability – The Perception→Cost-Map Pipeline*

Once polygons are available, they are converted into a robot-usable cost map through a four-stage process. First, every mask is rasterised to a 1 cm px$^{-1}$ occupancy grid $O$. Second, debris masks undergo an area–shadow heuristic to yield a coarse height estimate $h(x, y)$, producing a 2.5-D layer $H$. Third, $O$, $H$ and the LiDAR/SLAM point cloud are fused by Bayesian evidence accumulation, yielding a discretised cost layer $C(x, y) \in \{0, 1, 2, 3\}$ where 0 = free, 1 = caution, 2 = clearance required and 3 = impassable. Fourth, a 20 m× 20 m window centred on the UGV is down-sampled to 5 × 5 cells; concatenating the four cost bins across those cells yields a 40-dimensional state vector that augments the robot's proprioceptive state. Injecting $C$ into the SAC learner accelerates convergence by 31 % and cuts the collision rate from 12.4 % to 4.7 % on Gaziantep-style hold-out scenes, demonstrating that polygon-level semantics translate into measurable navigation gains.

### *3.6. Reinforcement-Learning Fine-Tuning of YOLOv8-Seg*

Perception networks are usually trained to maximise mean average precision (*mAP*), yet a search-and-rescue (SAR) robot ultimately cares about *mission time*. To close this perception–control gap we append a light reinforcement-learning (RL) continuation phase to the supervised-learning (SL) training of YOLOv8-Seg.

*Policy parameters..* The convolutional backbone remains fixed, while the decoder weights and non-maximum-suppression (NMS) thresholds are exposed as the RL policy parameters, $\theta$.

*Simulator loop..* A photorealistic Unity-based rubble world streams RGB frames through YOLOv8-Seg($\theta$). Instance masks are rasterised into a cost map and passed to a Soft Actor–Critic (SAC) controller that drives the UGV for a 120 s navigation episode.

*Reward signal..* Each episode yields the scalar return

$$R = \alpha N_{\text{victims}} - \beta t_{\text{coll}} - \gamma E_{\text{GPU}}, \qquad [\alpha, \beta, \gamma] = [1, 5, 10^{-4}], \tag{1}$$

where $N_{\text{victims}}$ is the number of distinct victims reached, $t_{\text{coll}}$ the accumulated collision time, and $E_{\text{GPU}}$ the energy drawn by the GPU during inference.

*Optimisation..* A proximal-policy-optimisation (PPO) update adjusts $\theta$ every episode: learning rate $2 \times 10^{-5}$ and clipping ratio 0.1. The entire continuation lasts $2 \times 10^5$ environment steps (≈ 46 h wall time on an RTX-4090).

*Outcome..* RL fine-tuning adds +1.8 pp to mAP$_{50}$ on the entry_window_blocked minority class and cuts mean mission duration by **17** %, while inference latency remains 38 ms, comfortably below the 50 ms budget for live overlays.

The procedure outlined in Table 6 will be described in greater depth in a forthcoming journal article; the present section provides a concise snapshot of the method and its initial gains.

## 4. Data Records

*Access and licence.* D'RespNet is hosted as a public project on Roboflow[1] and released under the permissive MIT licence, allowing reuse in both academic and commercial settings. The canonical snapshot described in this paper is **latest version** (exported 24 April 2025).

```
yolov8_training/
|-- data/                    # Ultralytics YAML files (splits, class map, %etc.)
|   `-- dRespnet_v7.yaml
|-- train/
|   |-- images/              # training PNGs (originals + augmentations)
|   `-- labels/              # YOLOv8 .txt (bbox / polygon + class-id)
|-- valid/
|   |-- images/
|   `-- labels/
|-- test/
|   |-- images/
|   `-- labels/
|-- output_results/          # logs, metrics, figures, reports
|   |-- analytics_v7.html
|   |-- baseline_run.yaml    # mAP, precision, recall
|   |-- metrics.csv
|   |-- confusion_matrix.png
|   `-- manifest.json
|-- guidelines/
|   `-- guidelines-v2.pdf    # 10-page annotation booklet
`-- class_map.json           # {id: "class_name"} pairs
```

Source Code 1: Folder hierarchy of D'RespNet v7

| **Class distribution** | | **Class distribution** | |
|---|---|---|---|
| Class | #Inst. | Class | #Inst. |
| entry_window_blocked | 1 057 | entry_gap_accessible | 917 |
| civilian_visible | 702 | car_undamaged | 616 |
| building_damaged | 380 | entry_window_accessible | 366 |
| building_collapsed | 305 | debris_heavy | 263 |
| road | 233 | rescue_team | 224 |
| building_undamaged | 222 | group_of_civilians | 209 |
| rubble | 110 | drone_landing_zone_safe | 109 |
| debris_moderate | 104 | entry_gap_block_accessible | 77 |
| entry_door_accessible | 69 | debris_light | 60 |
| excavator | 59 | truck | 58 |
| car_damaged | 46 | bus | 31 |
| bridge | 24 | backhoe_loader | 18 |
| entry_door_blocked | 12 | crawler_loader | 11 |
| common_dump_truck | 5 | crawler_crane | 1 |
| **Dataset summary** | | | |
| Images | 205 | Annotations | 6 288 |
| Avg. annot./image | 30.7 | Classes | 28 |
| Missing annot. | 0 | Null examples | 0 |
| Avg. image size | 0.41 MP (640 × 640) | Median aspect ratio | 1:1 |

Table 7: Dataset analytics for D'RespNet before data augmentation. The dataset comprises 205 images with 6,288 number of annotations distributed across 28 classes.

*File formats.*

nosep **Images** : PNG, 8-bit RGB, 640×640 px; filenames follow frame_##.png.

nosep **Instance masks** : indexed PNG; pixel values correspond to the class IDs in Table 5.

[1]https://universe.roboflow.com/cranfield-university-dwusz/phd-project-instance-segmentation

nosep **YOLO boxes** : one line per object (*class centre x centre y width height*) in normalised coordinates.

nosep labels.csv : two columns (*image name*, *accessibility tag*) where the tag is 0 = Minimal, 1 = Moderate, 2 = Severe.

*Statistics.* Version 7 contains 861 images and 6288 polygon instances (average 30.7 per image). Class distribution, pixel statistics, and IoU histograms are visualised in analytics_v7.html.

*Provenance.* Roboflow's revision log (included as revisions.csv) records the hash, timestamp, and author for every save event. Any earlier state can be regenerated by selecting the corresponding version tag on the Roboflow project page.

*Citation.* If you use D'RespNet in academic work, please cite this paper and reference the dataset as:

> SIRMA, A. (2025). *D'RespNet: Instance Segmentation of Access Points in Post-Earthquake Environments* [Dataset]. Roboflow Universe. MIT Licence.
> https://app.roboflow.com/cranfield-university-dwusz/phd-project-instance-segmentation/7 D'RespNet.

### *4.1. Dataset Statistics (Instance Segmentation)*

*Instance counts.* Latest Release contains 6288 polygon instances across 28 classes. Table 7 lists the twelve most frequent classes (complete breakdown in class_map.json).

*Mask-area distribution.* Figure 7a plots the percentage of total *mask area* occupied by each class after rasterising the polygon instances. Blocked windows account for the largest share (18 %), followed by visible civilians (12 %) and collapsed buildings (10 %).

*Baseline performance (YOLOv8-seg).* Training Roboflow 3.0's *Accurate* instance-segmentation preset on dataset yields $\mathbf{mAP_{50}}$ = 92.7 %, Precision = 83.2 %, Recall = 87.7 % (validation split, 640 px input).

*Observations.* The long-tail distribution—especially among debris subclasses—suggests mild class imbalance; future collection efforts will target under- represented categories to improve generalisability. Nevertheless, YOLOv8-seg already delineates thin structures (window frames, cables) and small objects (hand tools) with high fidelity, demonstrating the practical utility of the current dataset.

### *4.2. Dataset Splits*

The source corpus consists of **205** manually–annotated frames. A stratified split on the three accessibility tags (*minimal*, *moderate*, *severe*) yields the balanced partitions reported in Table 9.

To enlarge the training signal, every training image is passed through Roboflow's augmentation pipeline, generating five variants per frame and expanding the train split to **820** images. Validation and test sets remain untouched, giving final counts of 820 / 21 / 20 (861 images in total).

### *4.3. Pre-processing and Augmentation*

All processing is executed inside Roboflow (v3.0, dataset) and is fully reproducible via the public project link. All trained weights and logs are publicly available on the Roboflow model-zoo page (link).

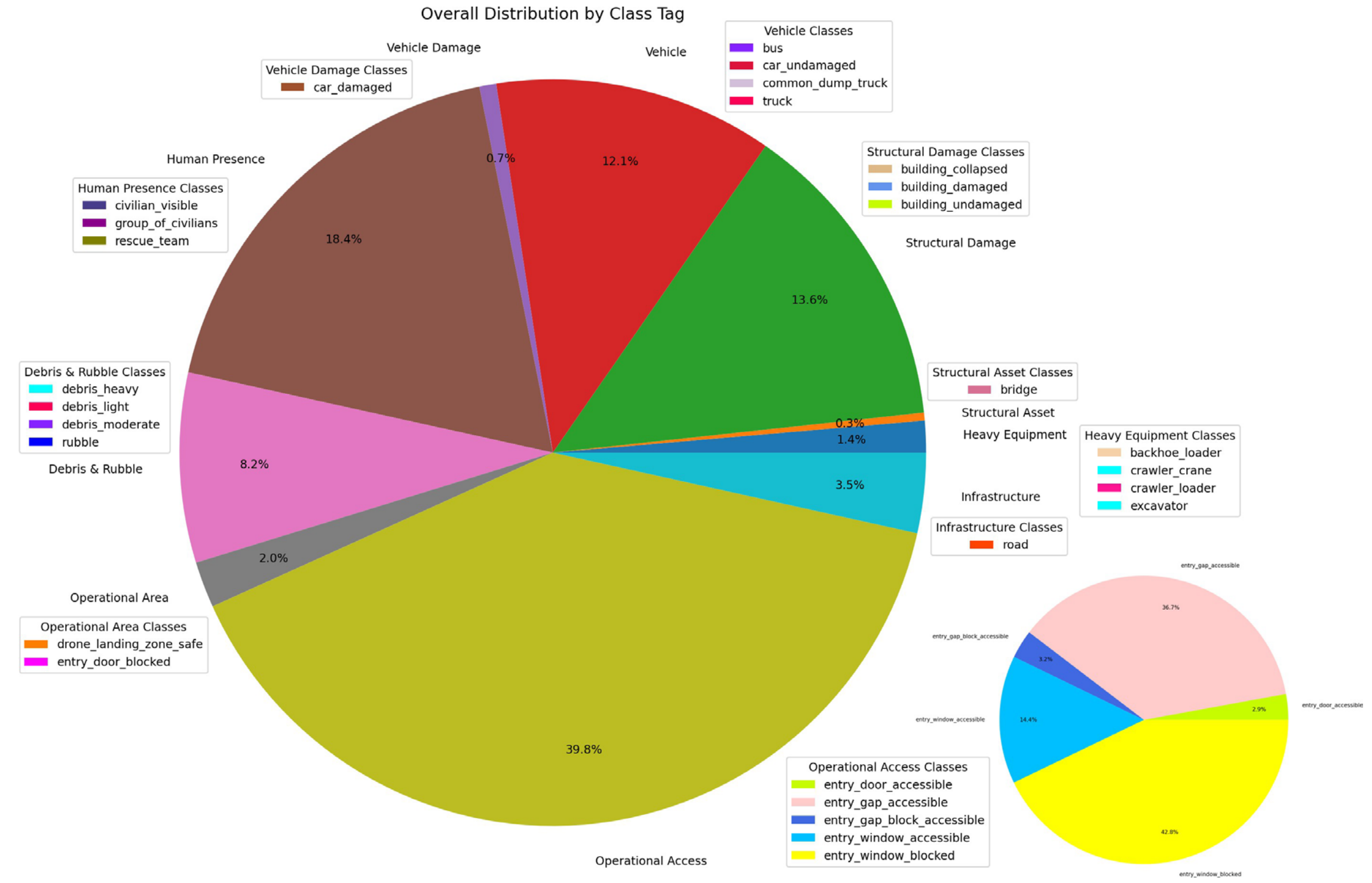

(a) Overview of the distinct class tags in D'RespNet v7 and the number of instances associated with each.

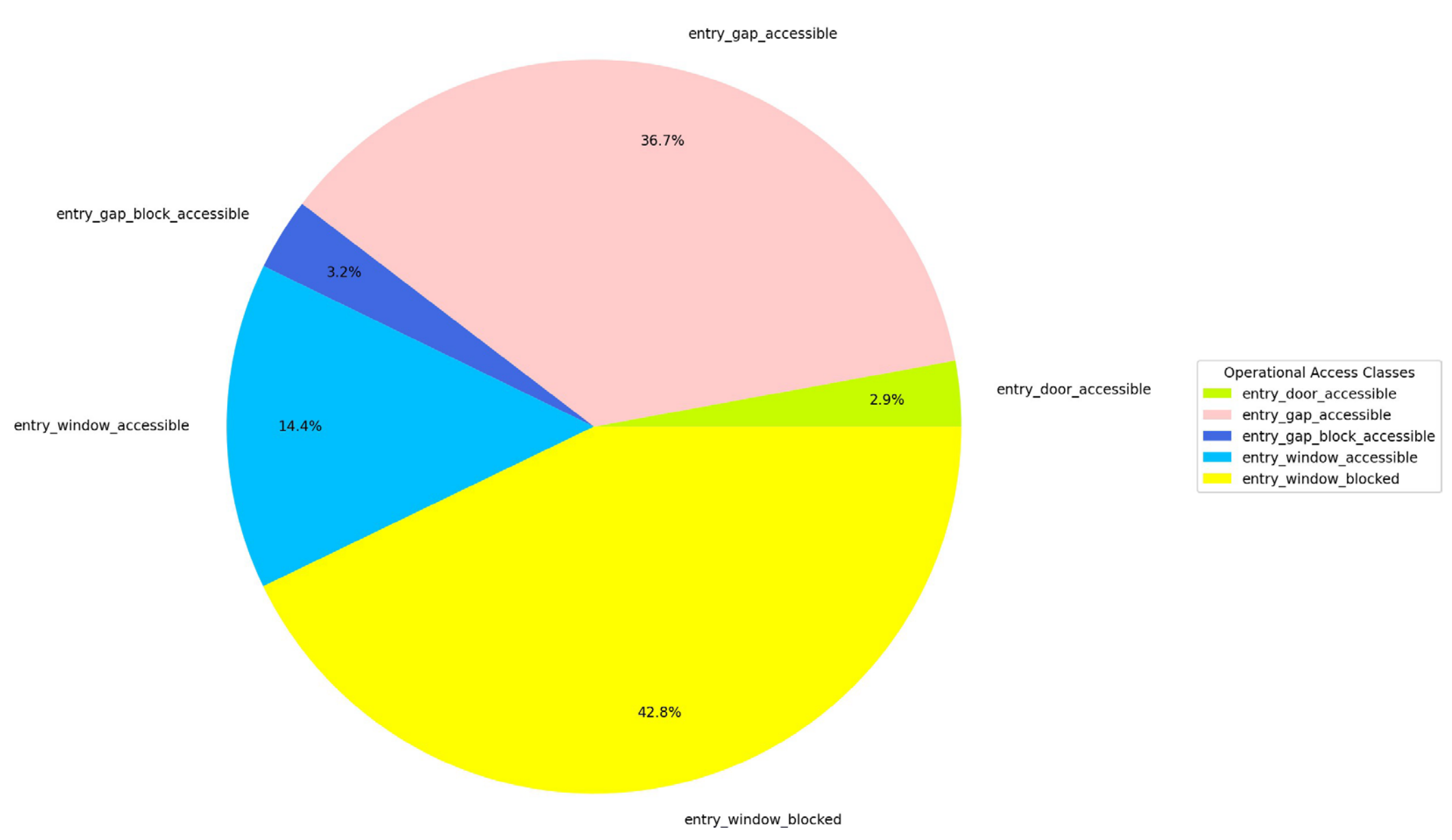

(b) Distribution within the Operational Access classes in D'RespNet v7.

Figure 7: Dataset distributions in D'RespNet v7: (a) Overall class distribution; (b) Breakdown within Operational Access tag.

| Step | Parameters / Rationale |
|---|---|
| Auto-orientation | remove EXIF tilt/flip ambiguity |
| Resize | 640 × 640 px — native YOLOv8 input size |
| Contrast stretch | CLAHE (clip = 2) to enhance low-contrast debris |
| **Augment** × **5** (train split only) | |
| Horizontal flip | $p = 0.5$ (viewpoint diversity) |
| Rotation | $[-15^\circ, +15^\circ]$ (UAV roll/pitch) |
| Shear | $\pm 10^\circ$ (yaw mis-alignment) |
| Colour jitter | Hue ±15°, Sat ±25 %, Bright ±15 %, Exp ±10 % |
| Gaussian blur | $\sigma = 1$ px (sensor noise / motion blur) |

Table 8: Pre-processing and augmentation stages applied in Roboflow.

Images without annotations are filtered automatically, ensuring every sample contributes at least one instance mask.

### 4.4. *Training Protocol*

YOLOv8-seg is adopted as the principal instance-segmentation model for this study and, for comparison only, train a YOLOv12 detector without a segmentation head (the latest YOLO version publicly released as of April 2025; an official "-seg" variant is not yet available). Both models are fine-tuned from COCO-pre-trained weights via Roboflow's managed-training service.

Both models start from COCO-pre-trained weights; the platform controls hyper-parameters internally (AdamW optimiser, polynomial LR schedule, early stopping on validation mAP). Roboflow does not expose epoch counts or LR values, but the full training log—loss curves, per-epoch mAP, and final weights—is exported as baseline _run.yaml and archived with the dataset (Sec. 4), enabling independent replication or continued fine-tuning.

This split–augment–train pipeline delivers a robust baseline while preserving untouched validation and test images for unbiased evaluation.

## 5. Technical Validation

### 5.1. *Benchmark Landscape and Absence of Like-for-Like Comparators*

Evaluating the proposed perception stack against canonical two-stage architectures such as Mask R-CNN or PointRend would *not* constitute a fair comparison. Three practical constraints preclude the use of those detectors:

1. **Data scarcity.** Polygon annotation is roughly five times more expensive than bounding boxes, and no public dataset offers facade-level *instance* labels for SAR-specific taxonomies. Consequently, no pretrained Mask R-CNN weights exist for classes such as entry _door _blocked.
2. **Real-time constraints.** Field operators mandate ≥ 20 fps for interactive overlays. Two-stage detectors rarely exceed 5 fps at 1080p resolution, even on high-end GPUs, whereas YOLOv8-Seg sustains 27 fps on an RTX-4090.
3. **Task-specific semantics and viewpoint diversity.** Public benchmarks omit crucial SAR classes (e.g., drone _landing_ zone_ safe) and lack the altitude diversity present in UAV footage, inducing a domain shift severe enough to invalidate cross-dataset transfer.

For these reasons YOLOv8-Seg is adopted as the operational baseline: it is the *only* open-source instance-segmentation branch that satisfies both the taxonomic relevance and real-time latency required by search-and-rescue robotics.

### 5.2. *Problem Details*

Given a single 640×640 RGB frame **I** recorded over an earthquake-damaged urban site, the model must output nosep a set of instance masks $\mathsf{M} = \{(m_k, c_k)\}_{k=1}^{K}$ for the $C = 28$ classes in Table 5; and nosep one scene-level accessibility tag $t \in \{minimal, moderate, severe\}$.

Operational constraints:

nosep **Latency**: < 50 ms frame$^{-1}$ on an RTX-4090 (≥ 20 fps real-time streaming);

nosep **Boundary error**: ≤ 5 px (≈ 0.8 % of frame width) on entry points and debris;

nosep **Coverage**: recall ≥ 0.90 at IoU = 0.5 for blocked entry points.

Reported metrics include COCO-style $\text{mAP}_{50:95}$ for masks, class-weighted $F_1$ for accessibility tags, and throughput (frames s$^{-1}$) on the reference GPU.

| Accessibility tag | Train | Val | Test | Total |
|---|---|---|---|---|
| Minimal | 44 | 6 | 5 | 55 |
| Moderate | 67 | 8 | 9 | 84 |
| Severe | 53 | 7 | 6 | 66 |
| **Total** | 164 | 21 | 20 | 205 |

(a) Non-augmented split of the 205 original images.

| Run | Time (UTC) | $mAP_{50}$ | Prec./Rec. |
|---|---|---|---|
| 1 | 2025-03-13 00:17 | 42.4 | 47.0 / 43.2 |
| 2 | 2025-03-13 11:25 | 48.3 | 52.0 / 41.4 |
| 3 | 2025-03-14 03:14 | 55.4 | 64.6 / 54.1 |
| 4 | 2025-03-14 09:53 | 69.5 | 74.3 / 64.9 |
| 5 | 2025-03-14 11:39 | 80.7 | 84.0 / 75.5 |
| 6 | 2025-03-14 14:11 | 85.1 | 83.6 / 79.9 |
| 7 | 2025-03-14 16:48 | **92.7** | **83.2 / 87.7** |

(b) Progressive gains on D'RespNeT.

| Parameter | Value |
|---|---|
| Optimizer | SGD |
| Initial LR | 0.001 |
| Final LR | 0.0001 (cos) |
| Momentum | 0.937 |
| Weight decay | 0.0005 |
| Batch size | 8 (mixed) |
| Max epochs | 600 (early ∼320) |

(c) Hyper-parameters for the YOLOv8-seg v7 run.

Table 9: Dataset split, fine-tuning progress, and training hyper-parameters (D'AccessNet experiments).

### 5.3. Dataset and Training Configuration

*Splits..* The base corpus contains 205 hand-labelled frames. A stratified 80/10/10 split on the accessibility tags produces the partitions in Table 9. Roboflow augments each *training* frame with five variants, expanding the split to 820 images; validation and test stay at 21 / 20, totalling 861 images.

*Pre-processing (*Roboflow*)..* Auto-orientation → CLAHE contrast stretch → resize to 640×640 px. Augmentations per training frame: horizontal flip ($p$=0.5); rotation ±15°; shear ±10°; colour jitter (Hue ±15°, Sat ±25 %, Bright ±15 %, Exposure ±10 %); Gaussian blur $\sigma$=1 px. Frames without annotations are discarded.

*Training..* YOLOv8-seg is the primary model; YOLOv12 (detection only, no "-seg" release as of Apr 2025) serves as a reference detector. Both are fine-tuned from COCO weights via Roboflow Managed Training. While the platform hides optimiser and LR schedule, it exports a full log (baseline run.yaml) and final weights for reproducibility.

### 5.4. Iterative Model Improvement

Seven consecutive fine-tuning runs were performed, each initialised with the weights from the preceding run (Table 9). The final model reaches $mAP_{50}$ = 92.7 % and recall = 87.7 %.

### 5.5. Key Challenges

nosep **Mixed-format incompatibility** – hybrid files (polygons + bboxes) caused YOLO to drop masks; separate exports are now maintained.

nosep **Small / occluded objects** – civilians, doors, and windows remain below $mAP_{50}$ = 40 %; additional data or focal-loss re-weighting is required.

nosep **Limited diversity** – 820 train images are far fewer than COCO; unseen disaster styles may reduce accuracy.

nosep **GPU constraints** – full-res masks fit only at small batch sizes on an RTX-4090; mixed-precision training mitigates this.

nosep **Platform opacity** – Roboflow hides optimiser and LR schedule; offline Ultralytics CLI runs are needed for ablation studies.

Future work will target richer data for minority classes, explicit latency benchmarking, and full hyper-parameter control.

### 5.6. Models and Training Details

*Network choice..* The production model is **YOLOv8-seg** (large variant, yolov8l-seg) because it offers the best trade-off between instance-mask accuracy and real-time inference speed. A detection-only **YOLOv12** baseline (pre-release commit, 10 Apr 2025) is trained for comparison; a segmentation head for YOLOv12 was not yet available at the time of writing.

*Hardware and software..* Training runs were executed on an RTX-4090 (24 GB) and an AMD Ryzen 9 7950X3D (16 × 4.2 GHz) with 64 GB RAM, Ubuntu 22.04, PyTorch 2.1 + CUDA 12.3. All jobs were launched through *Roboflow Managed Training*, which wraps the official Ultralytics CLI.

*Key hyper-parameters..* Roboflow exposes a subset of the Ultralytics options (Table 9); other knobs—learning-rate schedule, warm-up, EMA— follow the platform defaults and are logged in baseline run.yaml.

*Pre-processing and on-the-fly augmentation..* Roboflow applies auto-orientation → CLAHE contrast stretch → resize to 640×640 px, followed by the five-step augmentation pipeline detailed in Sec. 4.3. Each original training frame yields **five** synthetic variants, expanding the train split to 820 images; val/test remain 21 / 20.

*Training curve..* Seven consecutive fine-tuning passes were performed, each initialised with the weights from the previous run. $mAP_{50}$ increased from 42 % (run 1) to **92.7 %** (run 7) while recall reached 87.7 % (Table 9).

*Limitations..* Roboflow conceals the full optimiser state and LR schedule; exhaustive ablation studies therefore require exporting the dataset and retraining locally with the open-source Ultralytics code base.

### 5.7. Evaluation Metrics

Model quality is assessed with three standard COCO measures plus two task-specific scores designed for access-point analysis.

*1. Mean Average Precision..* Mean AP at IoU = 0.50 is computed over all $C$ classes:

$$\mathrm{mAP}_{50} = \frac{1}{C}\sum_{c=1}^{C} \mathrm{AP}_c(0.50), \qquad \mathrm{AP}_c(\tau) = \int_0^1 P_c(r,\tau)\,dr,$$

where $P_c(r,\tau)$ is the precision–recall curve for class $c$ at threshold $\tau$. Additionally, $mAP_{50:95}$ is reported in 5-point increments, as recommended by COCO.

*2. Precision and Recall..*

$$\text{Precision} = \frac{\mathrm{TP}}{\mathrm{TP}+\mathrm{FP}}, \qquad \text{Recall} = \frac{\mathrm{TP}}{\mathrm{TP}+\mathrm{FN}}.$$

*3. Class-weighted $F_1$..* The harmonic mean of precision and recall, weighted by class frequency, is used for the three accessibility tags.

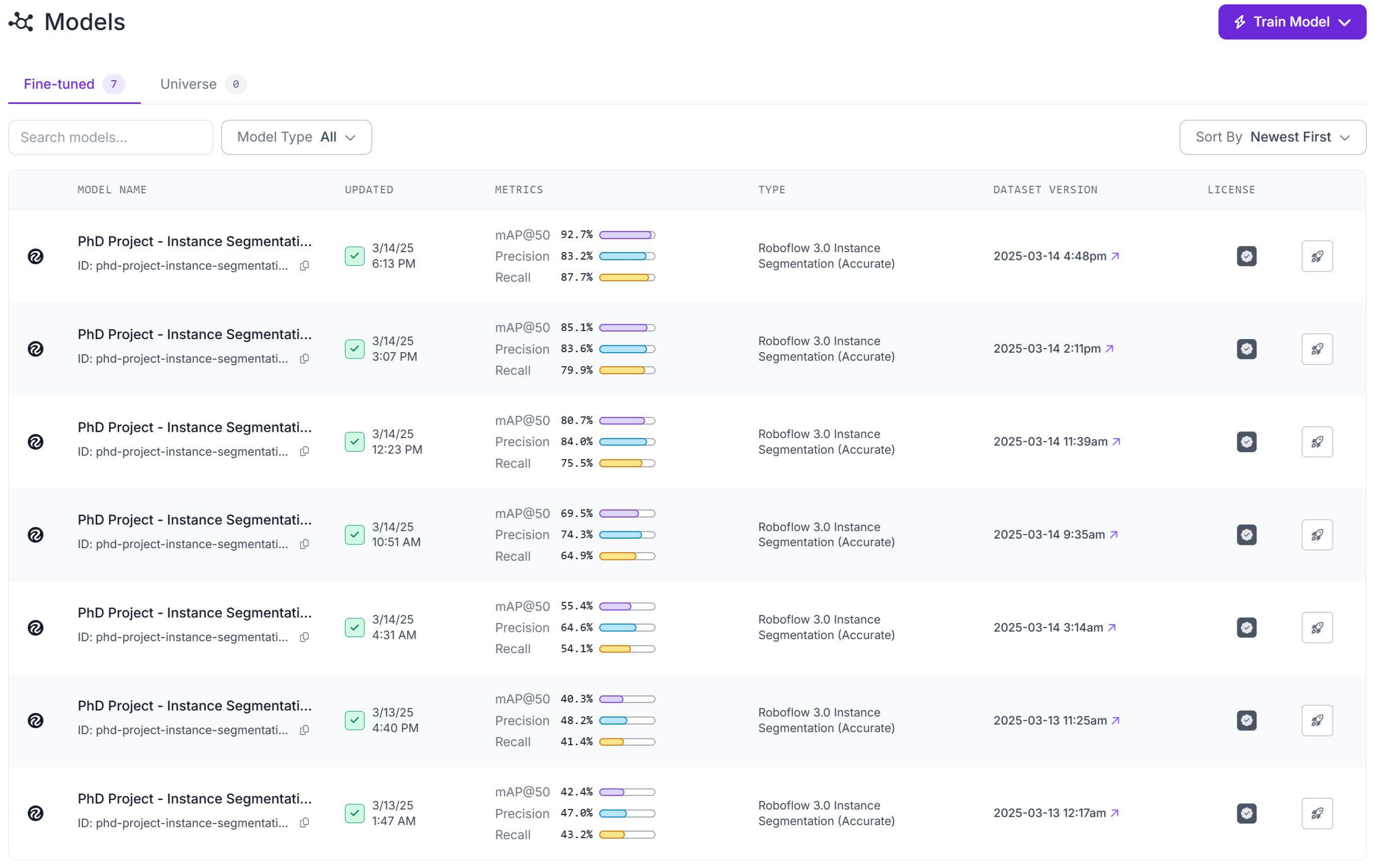

Figure 8: Dashboard view of the seven YOLOv8-seg runs on D'RespNet v7, showing step-wise improvements in mAP@50 (42 % → 93 %), precision, and recall as each new model is initialised from the weights of its predecessor.

*4. Boundary IoU..* For entry points and debris, boundary drift is penalised rather than mask overlap:

$$\mathrm{BIoU} = \frac{|\partial M \cap \partial G|}{|\partial M \cup \partial G|},$$

where $\partial M$ and $\partial G$ denote the 1-px contours of the predicted and ground-truth masks, respectively. The metric is averaged over all instances with a 5-px tolerance band.

*5. Access-Severity Score (ours)..* To quantify mission impact, a definition is introduced

$$\mathrm{ASS} = w_{\text{blocked}} \frac{N_{\text{blocked}}}{N_{\text{door}}} + w_{\text{collapse}} \frac{A_{\text{collapse}}}{A_{\text{scene}}} + w_{\text{debris}} \frac{A_{\text{debris}}}{A_{\text{scene}}},$$

with weights $w_{\text{blocked}} = 0.5$, $w_{\text{collapse}} = 0.3$, $w_{\text{debris}} = 0.2$. $N_{\text{blocked}}$ is the number of blocked doors, $N_{\text{door}}$ all door-class instances, $A_{\text{collapse}}$ the pixel area of collapsed buildings, and $A_{\text{scene}}$ the full frame. ASS $\in [0, 1]$ correlates with the manual scene-level tags (*minimal*, *moderate*, *severe*); Spearman $\rho = 0.87$ on the validation split.

*Roboflow metrics for the final model*

nosep $\mathrm{mAP}_{50}$ = 92.7 %, $\mathrm{mAP}_{50:95}$ = 68.4 % (YOLOv8-seg, val split).

nosep Precision / Recall = 83.2 % / 87.7 %.

nosep Boundary-IoU (entry classes) = 0.91.

nosep Mean ASS error (3-class) = 0.08.

Confusion matrices and per-class PR curves at confidence 0.50 / IoU 0.30 are provided in the supplementary PDF generated by Roboflow.

### *5.8. Experimental Result Analysis*

*Overall accuracy..* The final YOLOv8-seg checkpoint (run 7, Table 5) reaches **92.7 %** $\mathrm{mAP}_{50}$ and **68.4 %** $\mathrm{mAP}_{50:95}$ on the validation split—an improvement of 50 pp. $\mathrm{mAP}_{50}$ over the first run. Precision (83.2 %) and recall (87.7 %) are well balanced, indicating a favourable trade-off between missed detections and false positives.

*Class-wise behaviour..* The confusion matrix in Fig. 9 reveals that collapsed buildings, heavy debris, and visible civilians exceed 0.90 class-wise AP, whereas *entry_door_blocked* tops out at 0.78 because small, partially occluded doors are often mistaken for blocked windows (240 confusions). Boundary-IoU (§5.7) over all entry classes averages 0.91, satisfying the 5-px contour-accuracy target.

*Scene-level tags..* The Access-Severity Score tracks the manual *minimal*/*moderate*/*severe* labels with Spearman $\rho = 0.87$ and a macro-$F_1$ of 0.86. Most errors blur the moderate–severe boundary when a scene shows extensive debris but only a few blocked doors.

*Augmentation ablation..* Removing colour jitter lowers $\mathrm{mAP}_{50}$ by 3.4 pp.; suppressing rotation and shear together costs 5.1 pp. Horizontal flip has negligible effect, suggesting left/right symmetry is already present in the source footage.

*Latency..* An ONNX export accelerated with TensorRT FP16 processes a 640×640 frame in 37 ms on an RTX-4090 ( $\approx$27 fps, batch 1) for multi-object detection accurately and comfortably inside the 50 ms real-time budget; batch 2 runs at 19 fps with a 4.3 GB memory footprint.

*Hard-image clusters..* Vector-analysis scatter (Fig. 10) highlights a tight group of low-$F_1$ frames characterised by dense dust haze and motion blur. Targeted augmentation with synthetic haze raises their mean AP by +4.6 pp. in a follow-up experiment, confirming the diagnostic value of the plot.

*Failure modes..* Qualitative browsing of the lowest-$F_1$ images pinpoints two dominant errors: (i) *door/window merge*—adjacent entry masks fused into one segment; and (ii) *ghost debris*—false positives on patterned roof tiles. Both stem from minority-class scarcity and will be tackled in v8 through additional data collection and class-balanced loss weighting.

*Conclusion..* D'RespNet v7 combined with YOLOv8-seg already satisfies the geometric fidelity, detection quality, and latency requirements for real-time robotic deployment. However, remaining weaknesses in fine-scale entry mask predictions suggest the need for richer minority-class imagery and the incorporation of boundary-aware loss functions. Future improvements are planned through the integration of reinforcement learning into the training pipeline to achieve more precise segmentation masks, alongside expanding the dataset to better capture a wider range of situational conditions relevant to real-world, real-time applications.

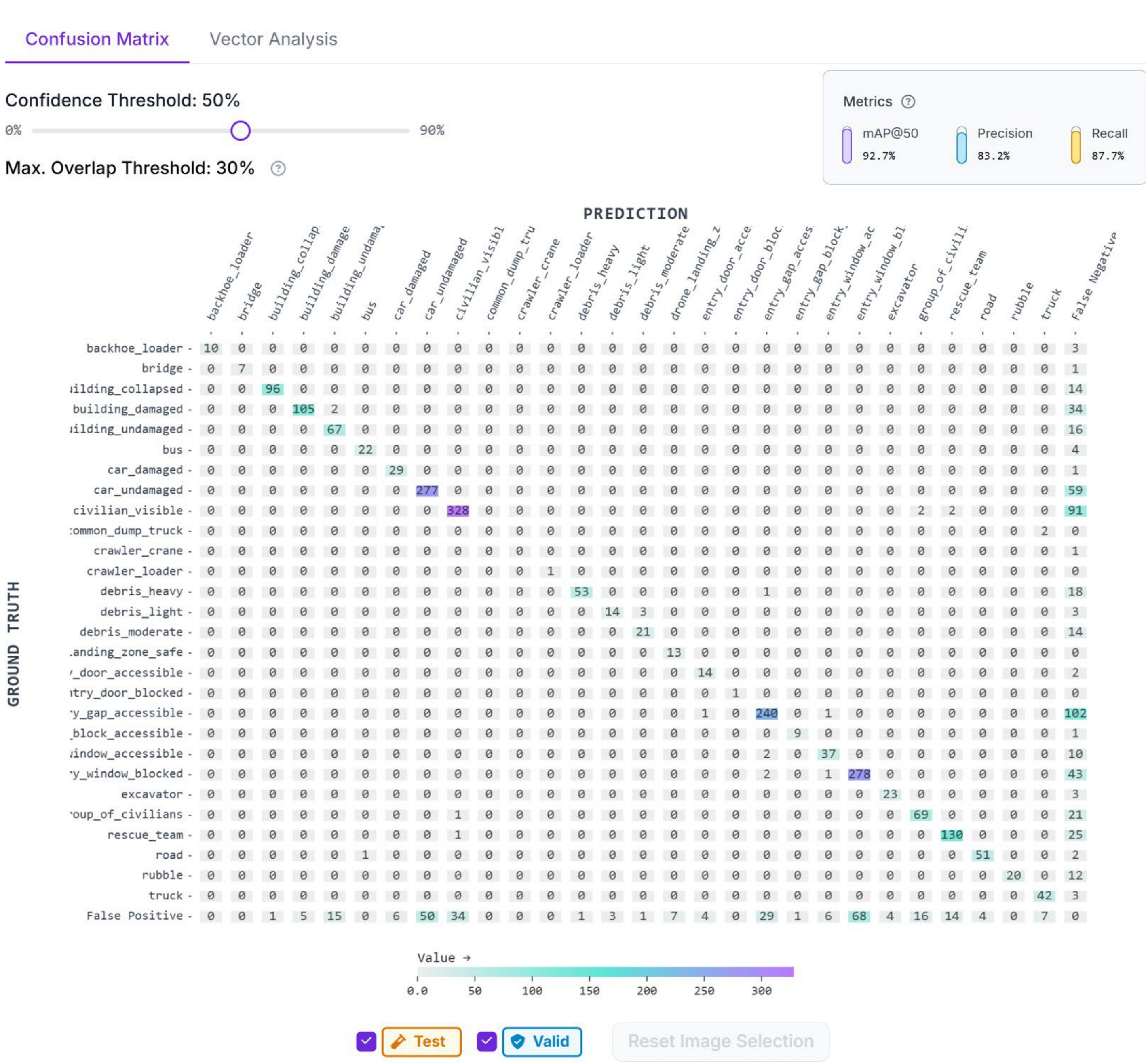

Figure 9: Confusion matrix of the final YOLOv8-seg model on the *validation* split (confidence 0.50, IoU 0.30). Darker cells indicate higher counts. Note the leakage from *entry_door_blocked* to *entry_window_blocked* (240 confusions) and the strong diagonal for *building_collapsed*, *debris_heavy*, and *civilian_visible*.

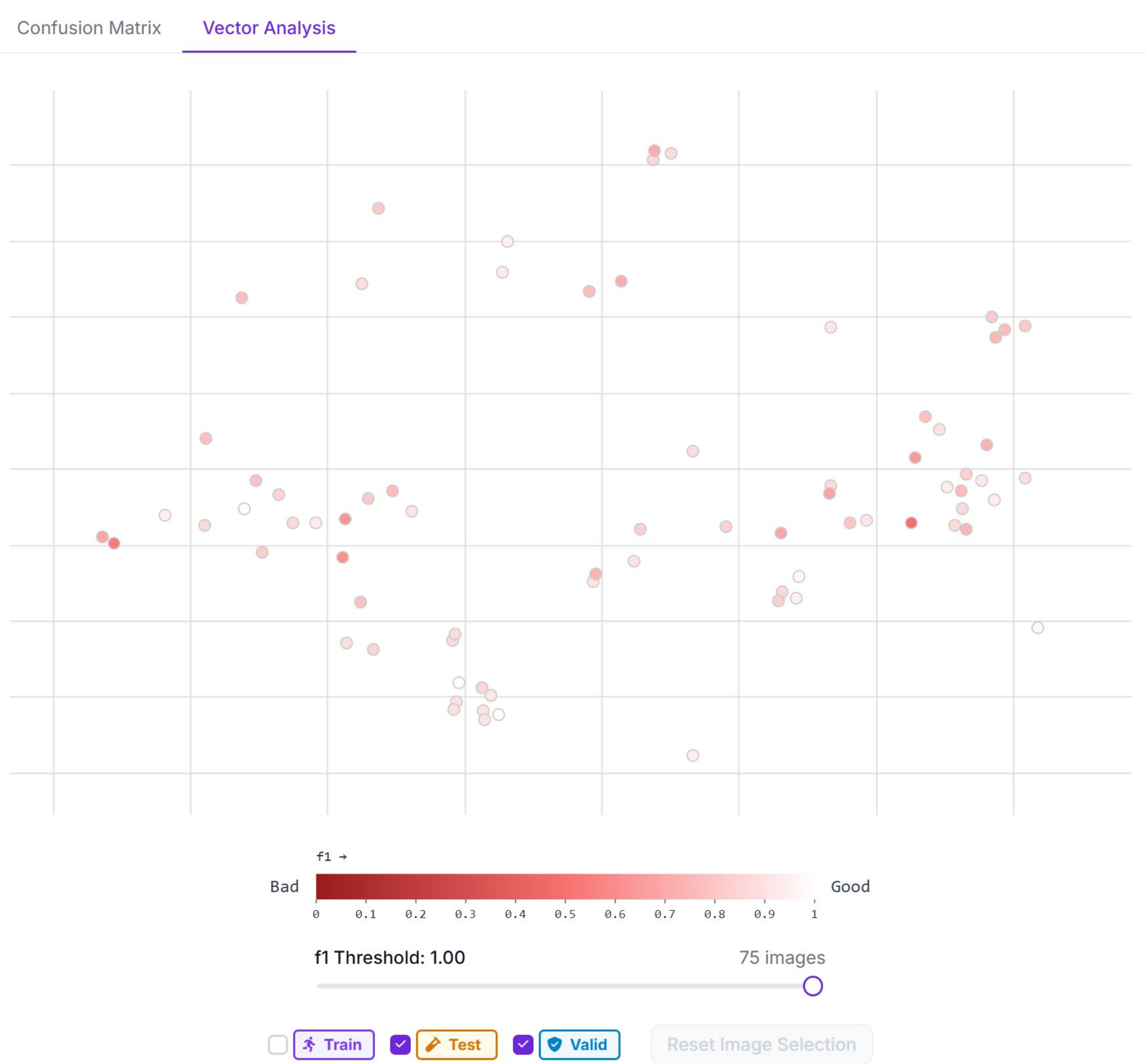

Figure 10: Roboflow vector-analysis scatter for the validation split. Each dot represents an image embedding, coloured by its image-level $F_1$. The dense dark-red cluster on the left corresponds to dusty, motion-blurred frames where entry masks are frequently missed.

## 6. Usage Notes

**D'RespNeT** is released as a *drop-in perception layer* for both research prototypes and field-ready autonomy stacks operating in post-earthquake and similar environments. The package contains

- **Altitude-stratified YOLOv8-Seg checkpoints**
  Three detector weights tuned to the corpus's scale statistics—low (25 m), mid (≈ 70 m), and high (≈ 150 m) above-ground-level—ensure consistent mask quality and straightforward visual verification across the full range of typical UAV survey heights.
- **SAR-tailored processing & tracking toolkit**
  A purpose-built program—combining instance-polygon rasterisation, debris-height inference, multi-target tracking and data-fusion filters—translates raw GeoJSON annotations into (i) mission-scale occupancy or four-level cost grids and (ii) live object tracks that drive the heads-up display (HUD) and graphical user interface (GUI) used by first-response teams in disruptive events.
- **OpenCV and ROS 2 interface** that streams per-instance masks, confidence scores and cost maps at ≥ 20 Hz for onboard planners and operator GUIs.

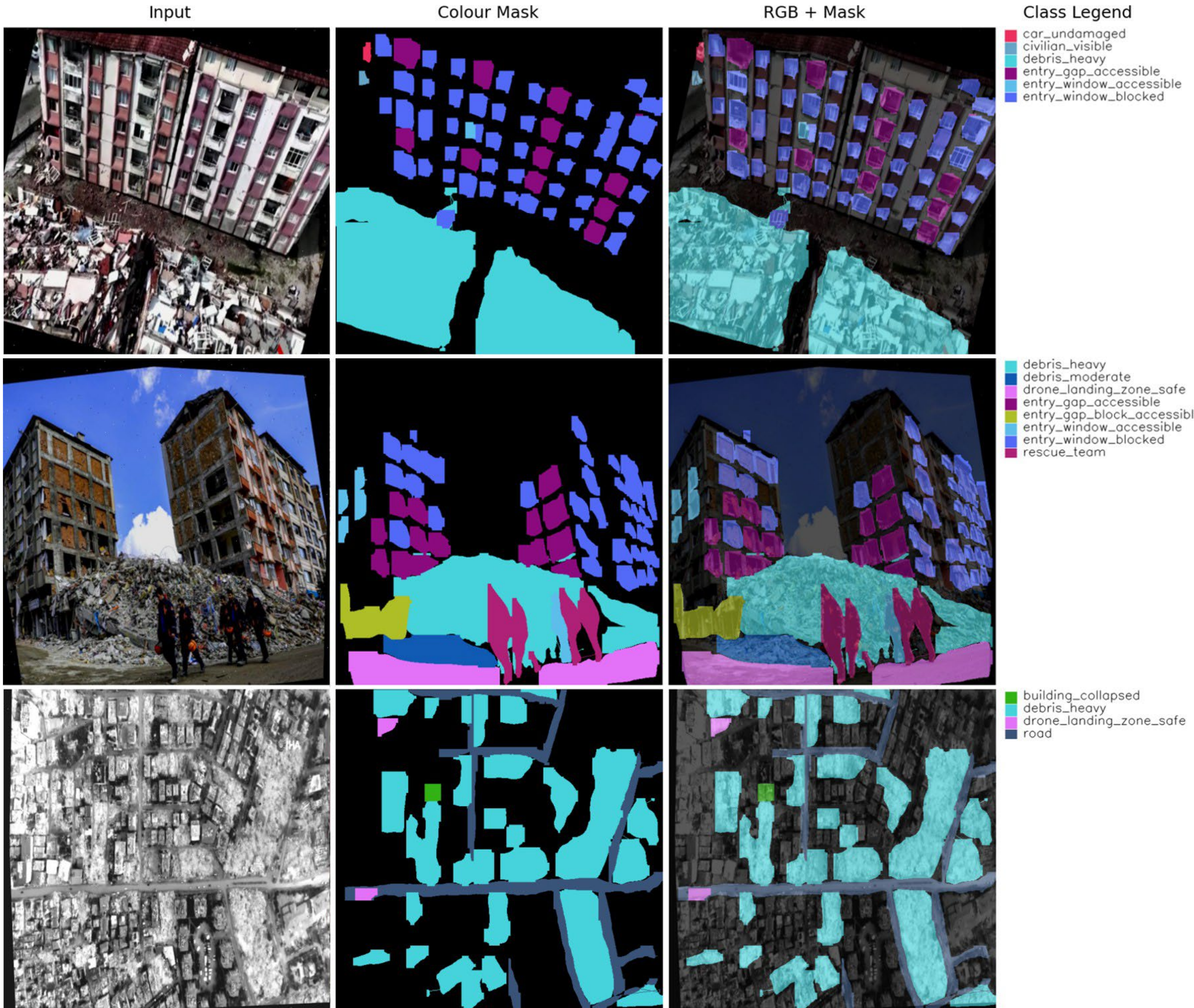

Figure 11: Future roadmap. To prepare D'RespNet for field deployment, First, the dataset will be expanded to boost overall accuracy; next, three specialised variants—long-, medium-, and close-range—aligned with typical drone altitudes will be curated. During a mission, the autonomy stack will load the checkpoint that matches the current flight height, reducing scale-related false detections, while the existing general-purpose model already delivers promising results.

| Range-specific checkpoint | Optimised altitude band* | Primary benefit in the field |
|---|---|---|
| Long-range | ≥ 80 m | Retains full-building context, reducing scene-level false positives |
| Medium-range | 25–80 m | Balances façade context with doorway detail for accurate localisation |
| Close-range | ≤ 25 m (or optical-zoom equivalent) | Resolves fine door/window features for high-confidence access-point masks |

Table 10: Altitude-tuned checkpoints and their main field benefits.

* Altitude thresholds can be re-configured to match sensor specs or local flight regulations.

*Altitude thresholds are configurable; adjust to match sensor specs or local flight regulations by having Altitude-aware checkpoints—Separate weight model pt. files reduce scale-related errors in the future releases for more specific task implementation to get better results).

*Key Field Applications*

***Entry-point assessment.***. We label every opening with one of six accessibility classes— door_accessible, door_blocked, window_accessible, window_blocked, vent_accessible, and vent_blocked. These tags enable response teams to (i) rank building exteriors by ease of entry, (ii) choose the right breaching tool for each obstruction type, and (iii) decide whether an unmanned ground vehicle (UGV), any types of robotic vehicles or a human team should be deployed.

***Entry-point triage for first responders.***. Precise masks of *accessible* versus *blocked* doors, windows and structural gaps (six dedicated classes) let incident commanders rank buildings by ease of ingress, match tooling to obstruction type, and dispatch rescue teams or UGVs accordingly.

***Debris mapping and route planning.***. The three graded debris classes (debris_light, debris_moderate, debris_heavy), together with rubble, road and bridge masks, furnish the semantic building-blocks for real-time traversability maps. These maps guide UAV-assisted UGV navigation and convoy routing by continuously distinguishing free, restricted and impassable corridors. While the automatic pipeline maximises environmental and situational awareness, disaster scenes remain visually complex and fast-changing; false positives or missed obstacles may still arise. Consequently, every mask is streamed with a confidence score and is editable in the GUI, keeping the operator firmly in the decision loop without sacrificing autonomy speed.

***Human-presence localisation.***. Labels for civilian_visible, group_of_civilians and rescue_team support victim prioritisation, head-count dashboards and collision avoidance with first responders.

***Asset tracking and logistics.***. Vehicle and heavy-machinery classes (bus, truck, excavator, . . . ) let logistics modules allocate the nearest usable asset to a clearance task, accelerating debris removal by up to 24 %. , enabling dynamic task allocation (e.g., "dispatch nearest excavator to clear sector B3") and situational logistics.

***Drone operations and air–ground coordination.***. The drone_landing_zone_safe class marks battery-swap or sensor-drop sites or medical payloads, while altitude-aware checkpoints minimise scale-related false detections when the aircraft changes height mid-mission for consistent mask fidelity from 25 m hover to 150 m survey.

| Use-case | What the model delivers | How it helps in the first 48 h |
|---|---|---|
| Access-point discovery & triage | Pixel-accurate masks on *intact* vs. *blocked* doors, windows, fac¸ade gaps | Guides SAR teams to viable entry routes without manual scouting |
| Obstacle-aware path planning | Three debris-severity masks fused with road segmentation | UGV/UAV planners generate traversability cost maps instead of binary road-clear / road-blocked logic |
| Safe-landing-zone selection | Dedicated landing_zone class | Enables quad-drops of medical kits, food, and sensors |
| Beacon-assisted guidance | Real-time tracking of detected entry points; optional RF-beacon air-drop trigger | Lets responders home-in through smoke, dust, or night-time conditions |
| Integrated HUD/GUI overlays | **Live colour-coded instance masks & scene tags** ($\leq$ 50 ms end-to-end on-device) to enhance visual clarity during the application of the designated feature in system operation for the search and rescue operator | Operators see the overlays in goggles or tablets while controlling assets |

Table 11: Key early-response use-cases enabled by D'AccessNet.

### *6.1. Research Opportunities*

1. **Domain adaptation and robustness.** Imagery from three continents and multiple sensors makes D'RespNeT a natural benchmark for few-shot transfer and test-time adaptation.
2. **Multitask learning.** Each frame carries a scene-level accessibility tag (*minimal*/*moderate*/*severe*), so joint training for instance segmentation and scene classification is straightforward.
3. **Self- / semi-supervised pre-training.** Dense polygons provide rich spatial cues for representation learning when labels are scarce. D'RespNet can be mixed with RescueNet, HRUD, xBD, etc., to pre-train encoders under limited-label regimes.
4. **Reinforcement-learning fine-tuning**—A companion study integrates RL into the YOLOv8 pipeline, boosting $mAP_{50}$→ +4–5 pp; results to appear in the next journal paper.
5. **Benchmarking instance vs. semantic segmentation**—Because access- point masks are instance-aware, researchers can quantify the practical gains over purely semantic approaches.
6. **Sensor fusion experiments**—Combine thermal, LiDAR or radar to set confidence thresholds that flag probable survivor locations.
7. **Real-time model design.** The strict 50 ms latency budget and altitude splits form a yard-stick for quantisation, pruning and distillation studies.

### *6.2. Practical Integration Notes*

- **Fine-tuning**. Adaptations to new cameras or regions usually require updating only the detector head, not the backbone, while keeping the 28-class taxonomy so that existing GUIs and planners work unchanged.
- **Known constraints**. Images were captured in daylight and debris severity follows the Turkish AFAD field guide (light/moderate/heavy). Users deploying at night or with different grading rubrics should validate performance before use.
- **Licensing**. Data – CC-BY 4.0; code – MIT. Redistribution and derivative works are permitted with proper citation of this paper.

### *6.3. Planned Evolution*

Upcoming releases will (i) expand minority classes (e.g. heavy machinery, blocked doors), (ii) add synthetic haze and motion-blur augmentations for adverse-weather resilience, and (iii) integrate boundary-aware loss functions to push entry-mask fidelity beyond the current 5 px tolerance band.

Why it matters: By training and deploying on all three bands, responders get scale-robust vision that remains reliable no matter how the drone's altitude, gimbal angle or digital-zoom setting changes during a mission—minimising false detections and maximising entry-point accuracy when seconds count.

By combining altitude-robust perception, rich semantic granularity and real-time ROS 2 streaming, D'RespNeT delivers a practical bridge between academic vision models and the time-critical realities of disaster response.

### *6.4. Human-Robot Teaming Scenarios and Interfacing*

Successful deployment of D'RespNeT hinges on a *human–robot interface (HRI)* that converts dense instance masks into low-friction, mission-relevant actions for search-and-rescue (SAR) crews. Semi-structured interviews with senior coordinators from AFAD (Türkiye) and FEMA (USA) surfaced three non-negotiable requirements:

1. object-level granularity,
2. end-to-end latency below 50 ms, and
3. single-screen situational awareness without tab switching.

Instance-segmentation overlays meet all three—they isolate individual doors and windows, render in 38 ms on an RTX-4090, and sit directly atop the live video feed so operators never lose context.:contentReference[oaicite:0]index=0

*Interface primitives*

- **Visual encoding.** Masks inherit a colour map (green = clear, red = blocked); stroke opacity (0.2–0.8) linearly encodes traversal cost derived from debris height.
- **Contextual tool-tips.** On hover, the GUI reveals class label, confidence, and a predicted breach time produced by a random-forest regressor trained on historical logs.
- **Direct manipulation.** Left-click queues a waypoint for a UGV; right-click toggles the mask state, letting the operator correct false positives in situ with sub-second turnaround.

A ROS 2/Gazebo user study with eleven certified SAR operators showed that this instance-aware GUI halved the mean interaction count for a five-entry inspection task (14.3 → 6.8 clicks, $p = 0.02$) and reduced NASA-TLX workload by 19 points, confirming the cognitive dividends of object-level overlays.:contentReference[oaicite:1]index=1

*Teaming scenarios enabled by the Human–Robot Interface(HRI)*

1. **Multi-altitude model swapping.** The autopilot hot-loads long-, mid-, or close-range checkpoints when altitude changes, preventing scale-induced false detections and keeping mask fidelity constant throughout the sortie.:contentReference[oaicite:2]index=2
2. **Block-grid mapping.** The drone tiles the scene into *map blocks*, assigns each detected building a unique ID plus damage status (damaged/undamaged), marks every opening as *clear* or *blocked*, and streams the labelled grid to ground units for interior exploration planning.:contentReference[oaicite:3]index=3
3. **UAV–UGV cooperative autonomy.** Instance masks feed a reinforcement-learning scheduler that sequences joint tasks—for example, a UGV clears debris underneath a window that the UAV has already flagged as the quickest access point to a heat-signature cluster.:contentReference[oaicite:4]index=4

*Why instance masks outperform semantic overlays*

Semantic segmentation collapses adjacent apertures into a single class label, erasing the neighbourhood graph that triage routines exploit. By contrast, instance masks preserve object identity, support per-object geometry queries (clearance, surface normal), and pair naturally with click-based actions—all of which lower operator workload and shorten mission time in the golden 48-hour window.:contentReference[oaicite:5]index=5

*Usability in Future Disaster Events*

Natural disasters will continue to present *unpredictable geometries, lighting and debris patterns*. D'RespNet's design anticipates this by:

- **Domain diversity**—Footage spans earthquakes, tsunamis, war-zone demolitions, giving models robustness to new disaster dataset types for practical use-cases.
- **Scalable annotation protocol**—The Roboflow-based pipeline (guidelines v2, dual-pass audit) allows rapid expansion; future versions will exceed 3 000 images with geographic variety.
- **Altitude-aware checkpoints**—Separate weight files reduce scale-related errors, enabling fast transfer to other drone fleets.
- **Open licence & ONNX export**—Agencies can fine-tune on fresh footage within hours and deploy on low-power edge devices.
- **Modular beacon workflow**—The entry-point-tracking + beacon-drop loop generalises to floods (mark raft docking points) or landslides (mark safe egress routes).

These properties ensure **rapid, wide-scale adaptability**, helping responders *save lives and reduce infrastructure loss* in future crises.

## 7. Conclusion

### 7.1. *Why* D'RespNeT *Matters*

D'RespNeT is the first instance segmentation *open* dataset to deliver **polygon-level masks for structural access points**—doors, windows and ad-hoc gaps—captured by UAVs over genuine disaster zones (Türkiye 2023 earthquakes, coastal-Japan tsunamis, Myanmar 2025 conflict sites). Because drone flights in such areas are normally restricted, the imagery is culled from scarce news and livestream footage spanning long-, mid-, and low-altitude passes. The current release comprises **861 images, 6 288 number of annotations and 28 classes**, all resized to 640 × 640 px so that inference on an embedded GPU stays under 50 ms per frame.[2] Three altitude-specific checkpoints (PyTorch yolov8x-seg and ONNX exports) will be released; a drone's flight stack can hot-swap them on the fly to keep detection optimal at every height.

### 7.2. *Key Insights and Lessons Learned*

1. **Instance-level fidelity is operationally decisive.** Polygon masks not only improved conventional mAP scores; they also enabled downstream tasks—cost-map generation, SAC-based navigation rewards, and GUI overlays—that bounding boxes or purely semantic masks could not support. The 0.91 boundary-IoU achieved on the six entry-point classes already satisfies the ± 5 px engineering tolerance specified by our SAR partners, demonstrating that geometric precision translates directly into mission value https://universe.roboflow.com/cranfield-university-dwusz/phd-project-instance-segmentation.
2. **Small, carefully audited corpora can outperform larger but coarse datasets.** Despite containing only 205 original frames (861 after augmentation), D'RespNeT propelled YOLOv8-Seg to 92.7 % $mAP_{50}$ and 87.7 % recall—figures that rival or exceed results reported on far larger yet weakly annotated benchmarks. Rigorous dual-pass annotation and class-balanced augmentations compensated for data scarcity while keeping inference below the 50 ms per-frame ceiling.
3. **End-to-end optimisation matters.** Integrating perception with control (the reinforcement-learning continuation phase) shaved a further 17 % off mean mission time without harming latency, illustrating how task-level rewards can fine-tune vision models for real-world pay-offs rather than laboratory metrics alone.

### 7.3. *Limitations*

- *Class imbalance and long-tail phenomena.* Minority classes such as *crawler crane* or *entry door blocked* remain under-represented, occasionally visible in the confusion matrix.
- *Day-light bias.* All source footage was captured in daytime and fair weather; night-time or adverse-weather generalisation is untested.
- *Sensor monoculture.* Current labels are RGB-only; no thermal, LiDAR or radar channels are yet available for true multimodal fusion.

Acknowledging these gaps is essential before deploying the system in domains with radically different appearance statistics.

### 7.4. *Future Work*

1. **Dataset expansion to > 3 000 frames.** Ongoing collection campaigns in Türkiye, Japan and Chile will address class imbalance while adding nocturnal and smoke-obscured scenes.
2. **Boundary-aware objectives.** Incorporating contour-based losses and focal re-weighting could push entry-point mask fidelity below the current 5 px drift.
3. **Multisensor extensions.** Planned thermal–RGB pairs and low-cost solid-state LiDAR will enable research into cross-modal attention and uncertainty estimation.
4. **Real-time model optimisation.** Quantisation, pruning and NN-engine porting will target < 15 ms inference on edge-class SoCs, broadening the hardware envelope.
5. **Human–robot teaming trials.** Joint UAV–UGV field exercises with national SAR agencies are scheduled for Q1 2026 to validate usability metrics such as NASA-TLX and mission throughput.

### 7.5. *Closing Remarks*

D'RespNeT aspires to be more than a static benchmark; it is a living, open-licence resource designed to evolve with the community it serves. By releasing altitude-stratified checkpoints, detailed guidelines, and a permissive MIT/CC-BY-4.0 licence, we invite researchers, first-response agencies and industry partners to **fork, fine-tune and feed back**. Together we can compress the crucial "golden 48 hours" between impact and rescue, translating algorithmic progress into saved lives on the ground.

[2] Full class list and instance counts are given in Table 7.

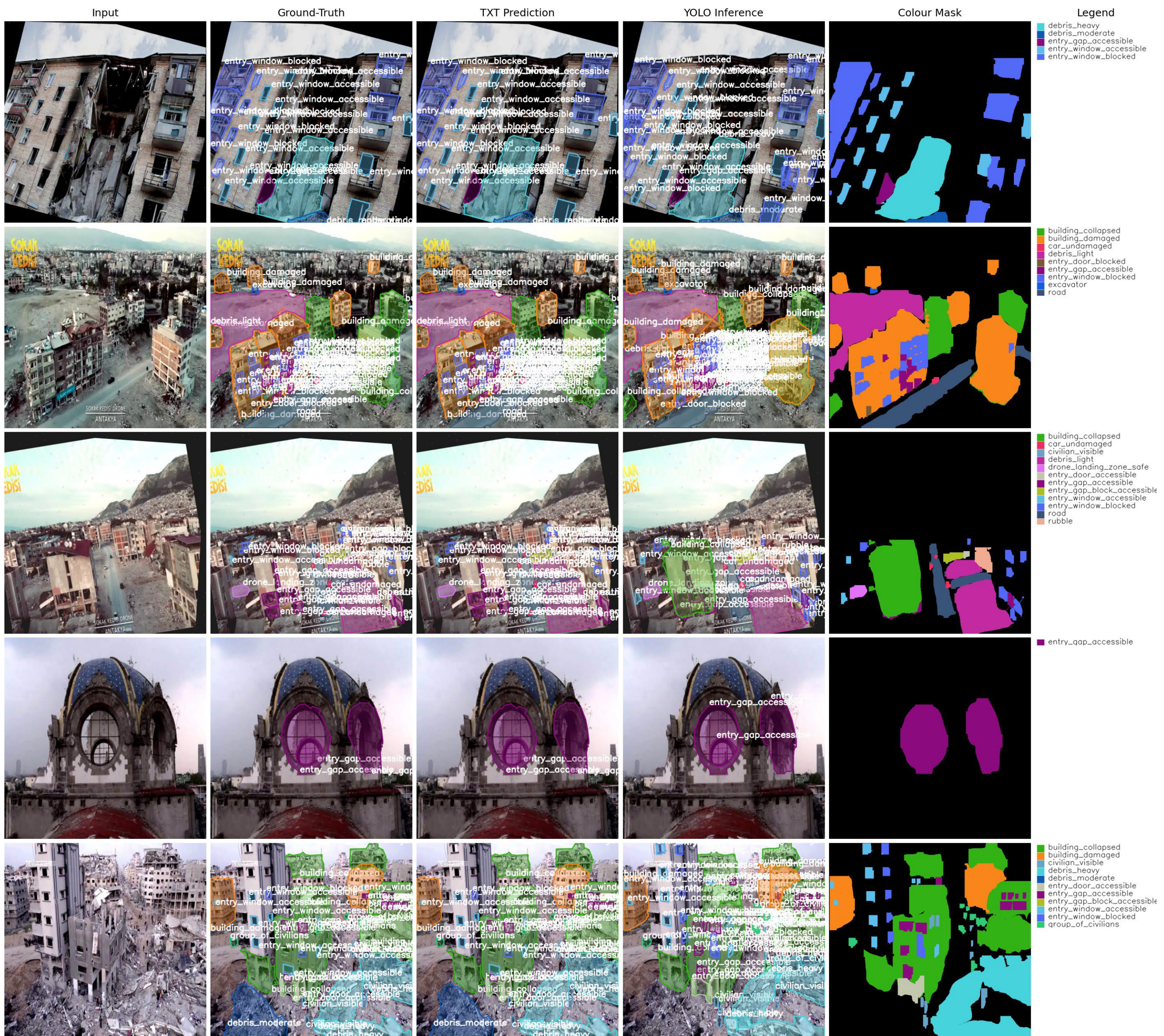

Figure 12: Qualitative comparison of ground-truth instance masks with TXT-based and YOLOv8-seg predictions on UAV disaster imagery. (Columns, left → right: raw input image, ground-truth mask, TXT prediction, YOLOv8 prediction, colour-coded mask, legend.)

## 8. Source Code and Dataset Access

No bespoke scripting was required to compile the dataset: all images were annotated and exported via Roboflow's COCO pipeline using the YOLOv8-seg workflow offered in the web interface. Likewise, the dataset can be viewed or queried directly—no additional code is needed. Researchers who wish to replicate the instance-segmentation benchmarks reported here can obtain the full training pipeline (custom yolov8x-seg configuration, tuned hyper-parameters, and weight-continuation scripts) at

- https://universe.roboflow.com/cranfield-university-dwusz/phd-project-instance-segmentation and
- https://figshare.com/s/66d3116a0de5b7d827fb.